%% file: main.tex
\pdfoutput=1
\documentclass[10pt,twocolumn,letterpaper]{article}

\usepackage[pagenumbers]{cvpr} 

\usepackage{graphicx}
\usepackage{amsmath}
\usepackage{amssymb}
\usepackage{booktabs}
\usepackage{makecell}
\usepackage{multirow}
\usepackage[mathscr]{eucal}
\usepackage[table]{xcolor}
\usepackage{algorithm}
\usepackage{algpseudocode}
\usepackage{epstopdf}

\usepackage{paralist}

\input{preamble}

\newcommand{\ourmethod}{RTracker}

\definecolor{cvprblue}{rgb}{0.21,0.49,0.74}
\usepackage[pagebackref=true,breaklinks=true,color links,citecolor=blue,linkcolor=blue,bookmarks=false]{hyperref}
\def\paperID{2962} 
\def\confName{CVPR}
\def\confYear{2024}
\newcommand*{\affaddr}[1]{#1} 
\newcommand*{\affmark}[1][*]{\textsuperscript{#1}}

\makeatletter
\def\thanks#1{\protected@xdef\@thanks{\@thanks\protect\footnotetext{#1}}}
\makeatother
\title{\ourmethod: Recoverable Tracking via PN Tree Structured Memory}

\begin{document}
\author{%
Yuqing Huang\affmark[1,2], Xin Li\affmark[2,$\ast$], Zikun Zhou\affmark[2], Yaowei Wang\affmark[2], Zhenyu He\affmark[1,$\ast$], and Ming-Hsuan Yang\affmark[3,4]\\
\affaddr{\affmark[1]Harbin Institute of Technology, Shenzhen}\quad
\affaddr{\affmark[2]Peng Cheng Laboratory}\\
\affaddr{\affmark[3]UC Merced}
\quad
\affaddr{\affmark[4]Yonsei University}\\
{\tt\small \{domaingreen2, xinlihitsz, zhouzikunhit, minghsuanyang\}@gmail.com, }\\
{\tt\small wangyw@pcl.ac.cn, zhenyuhe@hit.edu.cn}
%
}
\thanks{$\ast$ corresponding author}

\maketitle

\begin{abstract}
Existing tracking methods mainly focus on learning better target representation or developing more robust prediction models to improve tracking performance.
While tracking performance has significantly improved, the target loss issue occurs frequently due to tracking failures, complete occlusion, or out-of-view situations.
However, considerably less attention is paid to the self-recovery issue of tracking methods, which is crucial for practical applications.
To this end, we propose a recoverable tracking framework, \ourmethod, that uses a tree-structured memory to dynamically associate a tracker and a detector to enable self-recovery ability. 
Specifically, we propose a Positive-Negative Tree-structured memory to chronologically store and maintain positive and negative target samples.
Upon the PN tree memory, we develop corresponding walking rules for determining the state of the target and define a set of control flows to unite the tracker and the detector in different tracking scenarios.
Our core idea is to use the support samples of positive and negative target categories to establish a relative distance-based criterion for a reliable assessment of target loss.
The favorable performance in comparison against the state-of-the-art methods on numerous challenging benchmarks demonstrates the effectiveness of the proposed algorithm.
All the source code and trained models will be released at \url{ https://github.com/NorahGreen/RTracker}.

\end{abstract}

\section{Introduction}
\label{sec:intro}
%
Visual object tracking aims to estimate the location and extent of a target in a video sequence based on the bounding box annotation of the target given in the initial frame, which is a fundamental vision task with a wide range of applications, such as surveillance and autonomous navigation. 
The challenges in visual object tracking stem from the dramatic variations of the target (\eg rotation, deformation, and fast motion) and the various influences from the background (\eg occlusion, illumination variation, and out-of-view)~\cite{OTB2013}.
Existing tracking methods usually focus on learning robust target representation~\cite{TADT, TransT, SwinTrack, SimTrack, citetracker} or developing robust prediction models~\cite{DiMP, SiamRPN, SiamR-CNN, TrDiMP, OSTrack} to handle these tracking challenges and prevent target loss.
However, target loss, which can be caused by full occlusion, out-of-view, or tracking failure, is usually inevitable, especially during long-term tracking in complex real-world application scenarios.
Currently, considerably less attention is paid to the issue of how to recover tracking from target loss, which is becoming a bottleneck limiting the practical application of tracking algorithms.
%

\begin{figure}[t]
\begin{center}
\includegraphics[width=0.99\linewidth]{{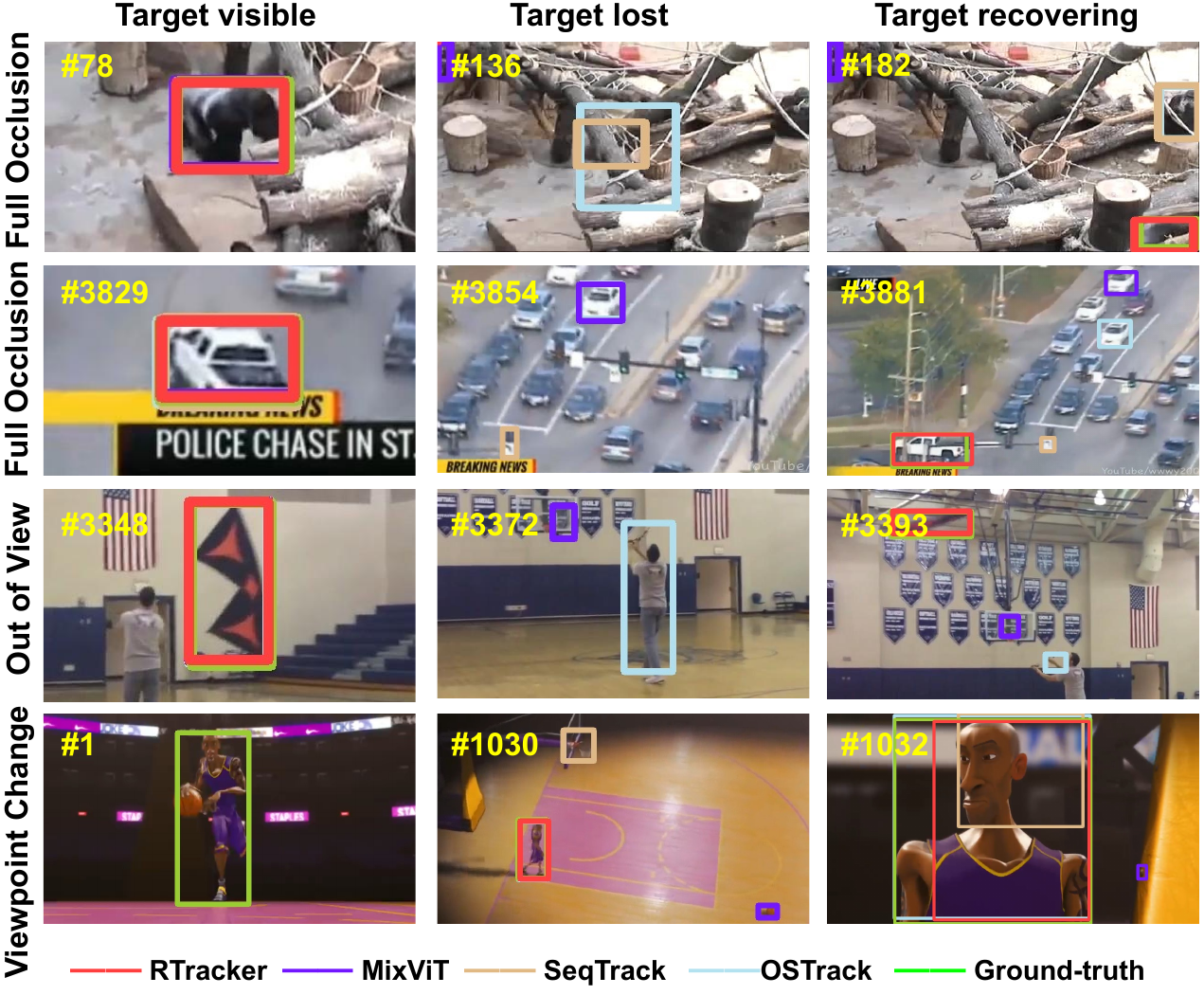}}
\end{center}
\vspace{-2mm}
\caption{\textbf{Performance on challenging sequences involving full occlusion, out-of-view, and viewpoint change.} 
The proposed RTracker can accurately re-track the targets in these sequences after their reappearance.
}
\label{fig:motivation}
\end{figure}

%
Enabling self-recovery capability in a tracking model is challenging since it first needs to accurately determine whether a target is present or absent and then provide the correct target information for re-initialization.
A few tracking methods~\cite{LCT} assess the presence or absence of a target based on the similarity between its current appearance and that of reference from the initial frame using a fixed threshold.
However, these may fail due to significant changes in target appearance over time and do not generalize well to different tracking scenarios.
%
%
Earlier trackers~\cite{TLD} explore bolstering the self-recovery of trackers by employing online adaptation, which can adjust the tracker in response to any deviation in the appearance of the target given at the beginning of tracking.
However, due to the substantial computational requirements and many training samples, performing effective online updates for deep learning tracking methods is impractical.

The key to addressing the above challenges lies in effectively modeling the continuously changing target over time, thus accurately determining the target presence/absence and re-initializing the tracking algorithm.
Instead of judging target states based on fixed thresholds, our core idea is to construct a relative measurement grounded in the positive and negative support vector of the target.
To enhance the reliability of target state evaluation, we should continuously refine the latest support vector to reflect the appearance changes and identify a closely matched complicated negative support vector for comparison.

To this end, we propose a recoverable tracking framework that dynamically associates detection and tracking via a specifically developed tree-structured memory to achieve tracking with self-recovery capability.
Specifically, we construct the Positive-Negative Tree Memory (PN tree), which archives appearance features relevant to the target, maintained according to temporal tracking results, as well as background information as negative samples that closely resemble the appearance of the target.
Upon the PN tree memory, we develop a series of walking rules to find the optimal support vector (target/background samples) for the current target and to ascertain its state (present or absent).
With the state identified, we apply pre-defined associating processes tailored to various tracking scenarios for achieving self-recovery for the tracker after target loss.
We conduct extensive experiments on a variety of challenging benchmarks, including VideoCube~\cite{videocube}, LaSOT~\cite{LaSOT}, LaSOT$_{ext}$~\cite{LaSOT_ext}, TNL2K~\cite{TNL2K}, and GOT-10k~\cite{GOT10k}.
The favorable performance against the state-of-the-art methods on all the benchmarks demonstrates the effectiveness of our proposed recoverable tracking algorithm.

We make the following contributions in this work:
\begin{compactitem}
    \item We propose a novel tracking framework capable of self-recovery for the tracker after target loss. 
    The proposed framework performs recoverable tracking guided by the established distinct control flows that integrate tracking and detection to manage the various states of tracking targets.
    \item We develop a Positive-Negative Tree Memory that stores and maintains positive and negative target samples over time.
    Additionally, we design a series of walking rules upon the PN tree to find the `support vectors' for determining the target present/absent.
    \item We achieve state-of-the-art performance across numerous tracking benchmarks.
    Extensive experiments, including ablation studies, are conducted to demonstrate the effectiveness of our proposed method and the effect of each component.
\end{compactitem}

\section{Related Work}
\label{sec: Related Work}
We discuss the closely related studies, including deep tracking methods, online adaptation trackers, and target search schemes for visual tracking.

\vspace{1mm}
\noindent\textbf{Deep tracking methods.} 
Deep tracking methods can be divided into Siamese-based and transformer-based categories based on the used backbone model.
%
%
Siamese-based trackers~\cite{SiameseFC, SiamRPN, SiamR-CNN, SimCLR} first compute the correlated features between the reference image and the test image, then predict the target state upon the correlated features.
%
%
In contrast, the transformer-based~\cite{TransT, OSTrack, blatter2023efficient,citetracker} trackers use successive transformer blocks to model the relationship between the reference and test images, thus achieving a more comprehensive correlation between them.
Siamese and transformer-based trackers only locate the tracked object in the current frame depending on the similarity between the reference and test images. 
This may result in tracking failure when the target is fully occluded or out of view.
Unlike the above trackers, our approach develops a tree-structured memory to dynamically model the target appearance by maintaining positive and negative target samples.

\vspace{1mm}
\noindent\textbf{Online adaptation trackers.} 
%
The main tracking challenges are caused by the variations of both the target and the background over time.
%
To combat that, several online adaptation techniques are developed, including online learning~\cite{TST, SPT}, template updating~\cite{CREST}, and memory networks~\cite{METrack, MemoryVOST}. 
%
Online learning tracking methods~\cite{DiMP, PrDiMP} continuously learn and model the targets throughout the tracking period, effectively adapting to changes in their appearance.
%
%
However, these methods may cause high computational costs due to the depth of neural networks. 
Other online adaptation methods~\cite{CREST, TATRack, SwinTrack} mitigate this by periodically re-initializing the tracking template based on a comparison to the reference with a fixed threshold. 
Besides, several methods~\cite{METrack, STMTrack} utilize dynamic memory networks to merge the initial template with historical tracking information for enhanced adaptability.
%
However, previous online adaptation tracking methods depend on fixed thresholds for updates, which does not generalize well to different sequences.
Our proposed method improves the accuracy in determining the target state by comparing the relative distances between positive and negative samples stored in a PN tree memory structure, which adaptively saves target information and offers a more reliable adaptation.

\vspace{1mm}
\noindent\textbf{Tracking aided by detection.} 
To address the problem of target loss, several tracking methods~\cite{TLD, iccv19_SPLT, RANSAC} explore a global detector to aid tracking by detecting the target after its reappearance.
%
%
%
Kalal~\etal propose a tracking-learning-detection algorithm (TLD)~\cite{TLD} for handling target loss, which utilizes bidirectional optical flow matching as the local tracker and an online detector based on ensemble learning for global detection. 
TLD also employs a nearest neighbor classifier to determine the target state and then uses a learning module to associate the tracker and detector based on the target state.
Ma~\etal~\cite{LCT} use a discriminative correlation filter based on the histogram of orientation gradient features for local tracking and implement global detection using an online random fern classifier. 
Unlike these methods that mainly relied on a classifier to associate trackers and detectors, our proposed method includes constructing relative measurements based on positive and negative target samples to evaluate the target state and associate trackers and detectors more reliably.

\section{Proposed Algorithm}
\begin{figure*}[t]
\centering
\includegraphics[width=0.98\linewidth]{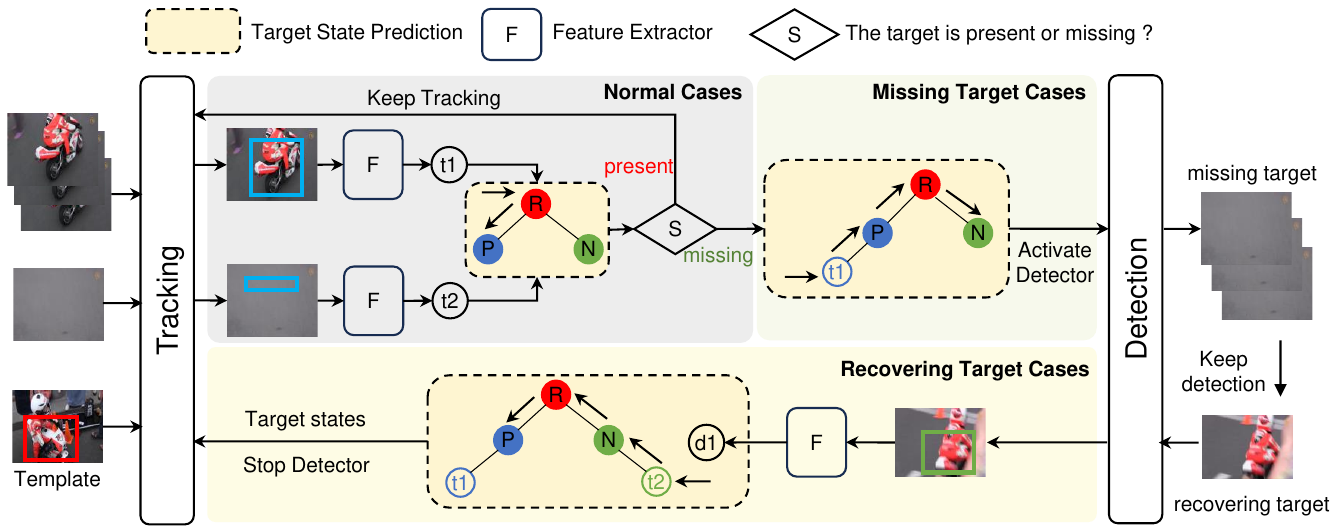}
 \caption{\textbf{Overall flow of the proposed algorithm.}
 Our proposed tracking method dynamically associates the tracker and detector based on the target state. 
 It contains three associating processes: 1) normal case flow, which validates the target state normal, utilizing only the tracker for tracking; 2) target missing flow, which confirms the target as lost, activating the detector for global searching; 3) target recovering flow, which detecting until the lost target recovering, the detector stopped and reactivating the tracker.
 }
 \vspace{-2mm}
\label{Fig:framework}
\end{figure*}

The goal of our method is to dynamically associate a tracker with a detector to achieve recoverable tracking based on target states (\ie presence or absence) over time, where the tracker accounts for precise target localization in successive frames and the detector accounts for the global searching.
To this end, we propose a Positive-Negative Tree (PN tree) structured memory equipped with a set of walking rules to achieve a reliable determination of the target state.
We then associate the tracker and detector using the pre-defined control flows based on the target states to perform tracking.
Figure~\ref{Fig:framework} depicts the overall flow of the proposed method, which includes the control flows for the normal case, the target missing case, and the target recovering case.




\subsection{PN Tree Structured Memory}

\begin{figure}[t]
    \centering
    \begin{subfigure}[b]{0.49\linewidth}
        \includegraphics[scale=0.7]{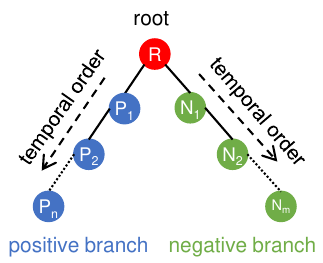}
        \caption{Structure of the PN tree}
        \label{fig:a}
    \end{subfigure}
    \begin{subfigure}[b]{0.49\linewidth}
        \includegraphics[scale=0.7]{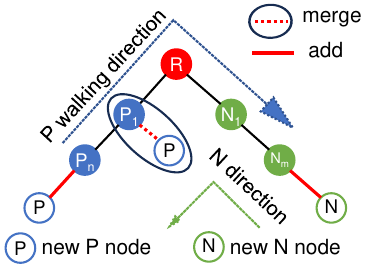}
        \caption{Updating of the PN tree}
        \label{fig:b}
    \end{subfigure}
\caption{\textbf{Definition of the PN tree.} 
 R, P, and N denote the root node, positive nodes, and negative nodes, respectively.
 New nodes are added to the tree through two ways: addition and merging. 
 The walking paths are used for finding the support samples for target state determination.
 }
 \vspace{-4mm}
\label{Fig:tree}
\end{figure}

The primary issue in self-recovery tracking is to determine whether the target is present or missing in a test frame.
Instead of judging target states based on fixed thresholds, our core idea is to construct a relative measurement based on positive and negative target samples to enable a more reliable assessment of the target states.
The Positive-Negative Tree is constructed to maintain the `support vectors' for positive and negative target samples over time, which is akin to the SVM algorithm.
In addition, we propose a set of walking rules for exploiting the PN tree memory to determine the current target state.
Figure~\ref{Fig:tree} depicts the structure and updating operation of the proposed PN tree.

\vspace{2mm}
\noindent\textbf{PN tree definition.}
%
We define the PN tree as a specialized variant of the binary tree, characterized by a root node and two distinct branches: the positive and the negative.
The root node stores the features of the target template given in the initial frame, and its two branches maintain the target-relevant information as the positive branch and the target-irrelevant samples as the negative one.
Each node in the tree structure stores the features of a representative target sample at a specific stage extracted by a similarity perception model~\cite{dreamsim}.

\vspace{2mm}
\noindent\textbf{Operations of the PN tree.}
To meet the target modeling requirements of visual tracking, we define the initialization, update, and deletion operations of the PN tree.
\textbf{Initialization.} The root and the first positive child node of the PN tree are initialized using the features of the target template given in the initial frame. 
Simultaneously, the initial negative node is constructed with target-irrelevant information (\ie features of a background sample).
%
\textbf{Update.} 
For an efficient PN tree memory, we adopt two different sample update strategies based on whether the target sample contains a new target appearance.
%
For the updated sample without a new target appearance, 
we merge the new node with the existing child node (excluding the root) via Equation~\ref{merge}. 
After merging, we move the node to the deepest node in the positive branch.
For the updated sample with a new target appearance, we directly append the new node as the deepest positive node in the PN tree.
The merging process is formalized as follows:
\begin{equation}
\begin{aligned}
\label{merge}
    F_{new} &= (F_x + F_{old} \times  N )/(N + 1),
\end{aligned}
\end{equation}
where $F_{new}$ and $F_{old}$ are the features of the node before and after the merge operation, N denotes the number of updates applied to this node, and $F_x$ is the feature of the new sample to be updated.
For the negative branch, we add the new negative sample as the deepest negative node.
Hence, within the PN tree, the depth of a node serves as an indicator of its temporal updated order, where nodes situated at greater depths correspond to more recent updating.
\textbf{Deletion.} When a branch exceeds N nodes, the earliest node is deleted.
This study sets N to 10, balancing memory efficiency and tracking accuracy.

\vspace{2mm}
\noindent\textbf{Walking rules of the PN tree.}
Our key idea is to identify the type (positive or negative) of the test node based on its similarity to the nodes in the PN tree.
To this end, we define the walking rules with three elements: walking operation, stopping condition, and walking path.
\textbf{Walking operation.}
For each node walked by, we compute its similarity against the test node using the cosine similarity defined as:
\begin{align}
\label{fine_feature_compare} S &= cos(F_x,F_{node}),
\end{align}
where $F_x$ and $F_{node}$ denote the test node and the candidate node, respectively.
\textbf{Walking path.}
%
We define a positive path to identify whether the node transitions from a positive label to a negative label and a negative path to identify transitions from a negative to a positive label.
The positive path goes from the leaf nodes of the positive branch up to the root node and then descends from the root node to the leaf nodes of the negative branch.
The negative path goes in reverse.
\textbf{Stopping conditions.}
For the positive cases, the walking process stops when a node more similar to the test node than the root is encountered after walking through both positive and negative branches.
%
The label of the test node is then determined based on the label of the branch where the walking stops.
If such a node is not encountered, the label of the test node is determined as positive.
For the negative path, the walking process continues until the entire route is completed, and the label of the test node is predicted as the label of the node with the highest similarity encountered along the entire path.
%
%
The walking rules guarantee a reliable classification of nodes by employing relative similarity measurements between positive and negative nodes.

\subsection{Associating Tracking and Detection}
\label{AssTD}
%
%
%
%
In this section, we use the proposed PN tree to predict the state of the target and dynamically associate the tracker and detector based on the target states.
%
%
%
According to different tracking scenarios, we define three processes, including normal case flow, target missing flow, and target recovery flow to control the operation of the tracker and the detector, which is shown in Figure~\ref{Fig:framework}.

\vspace{1mm}
\noindent\textbf{Target State Prediction.}
We use the PN tree, incorporating defined operations and walking rules, to model the target appearance and identify the target state.
The PN tree models the tracking target by forming a node of the target
using the extracted features based on the tracking or detection results.
We then identify the type of the newly formed node and update it into the PN tree using the walking rules and update operation, respectively.
A positive node signifies successful tracking (normal case), while a negative node indicates target loss (missing target case).
We then handle these cases using the following processes.

\vspace{1mm}
\noindent\textbf{Normal case.}
As shown in the top line of Figure~\ref{Fig:framework}, we only exploit the tracker to follow the target object.
Figure~\ref{Fig:framework} shows the process of the normal case in the top-left part.
For each frame, we use the tracker to predict the location of the target object and use the PN tree to identify the state of the target.
If the target is present, we continue tracking and update the PN tree memory with the new tracking results.

\vspace{1mm}
\noindent\textbf{Missing target case.}
The tracker stops if the state prediction infers the target is missing within the search region, as shown in the top-right of Figure~\ref{Fig:framework}.
We then activate the detector for a global search at this point. 
Concurrently, the latest tracking result, classified as a negative sample, is added to the PN tree memory as a new negative node. 
This step enriches the PN tree memory with the most recent information and improves state prediction by incorporating examples of failed tracking. 
%
%

\vspace{1mm}
\noindent\textbf{Recovering target case.}
For the target recovering flow, shown in the bottom part of Figure~\ref{Fig:framework}, the detection task continues until the target is detected in a frame.
%
%
%
When the target reappears, the labels of the nodes corresponding to the successive frames will change from negative to positive.
Therefore, we select the negative path to traverse the PN tree and determine the state of the target. 
%
%
Upon confirming the recovery of the target, we add the new target into the PN tree memory.
Simultaneously, we deactivate the detector, activate the tracker, and provide the tracker with the location of the current target for the following tracking.
%

\subsection{Recoverable Tracking}
We illustrate the proposed recoverable tracking pipeline in Figure~\ref{Fig:framework} and provide the pseudo-code in Algorithm~\ref{algorithm:RTrack}.
In the initial frame, we initialize the tracker, detector, and PN tree memory using the given target exemplar.  
For every following frame, we perform target state prediction and then handle different cases using the corresponding processes defined in Section~\ref{AssTD} to perform tracking, detection, and updating dynamically.
\begin{algorithm}
\footnotesize
\caption{Recovering Tracking Algorithm}
\label{algorithm:RTrack}
\begin{algorithmic}[1]
\Require Video $V = (I_0, I_1, \ldots, I_t)$, Tracker $T$, Detector $D$, Reference $R$
\State $M \gets \Call{InitMem}{R}$ \Comment{Init PN tree M with R}
\State $S \gets \text{1}$ \% Init target State S as present
\State Initialize $T$ and $D$ with $R$
\While{frame $I_i$ in $V$ }
    \State $Results \gets T(R, I_i)$
    \State $S \gets \Call{Walking}{M, Results}$
    \If{$S = 1$}
        \State $M \gets \Call{Update}{M, Results}$
        \State \textbf{continue}
    \Else
        \State $M \gets \Call{Update}{M, Results}$
        \While{$S = 0$} \% Target is missing
            \State $Results \gets D(I_i)$
            \State $S \gets \Call{Walking}{M, Results}$
            \If{$S = 0$}
                \State $i \gets i + 1$
            \Else
                \State $M \gets \Call{Update}{M, Results}$
                \State $T \gets \Call{ReInit}{Results}$ \Comment{Reinit tracker}
            \EndIf
        \EndWhile
    \EndIf
    \State $i \gets i + 1$
\EndWhile
\vspace{-2mm}
\end{algorithmic}
\end{algorithm}

\section{Experiments}
In this section, we present the experimental results of our proposed RTracker. 
We first compare the overall performance on five large-scale challenging tracking benchmarks against the state-of-the-art trackers.  
We then conduct a comprehensive ablation study to analyze the contribution of each component.
A recovery ability evaluation is performed to demonstrate the effectiveness of our tracking method in successfully recovering the tracking target once it has been lost.
Finally, the visualized results on several challenging sequences are provided to present an exhaustive qualitative analysis. 
More detailed results and experimental settings can be found in the supplemental materials.

\subsection{Implementation Details}
Our experiments are conducted using 4 NVIDIA Tesla V100 GPUs.
We employ the MixViT-L (ConvMAE)~\cite{cui2023mixformer} as the base tracker in conjunction with the MITS~\cite{mits} model as the based detector. 
For the tracking model, we crop the search image that is 4.5 times the area of the target box from the test frame and resize it to a resolution of $384 \times 384$ pixels.
The template is cropped as twice that of the target box and has a resolution of $192 \times 192$.
We train a similarity perception model following the setting of dreamsim~\cite{dreamsim} on the LaSOT and NIGHTS~\cite{dreamsim} datasets as the feature extractor in the PN tree.

\subsection{State-of-the-Art Comparison}
We compare our tracker with the state-of-the-art tracking algorithms on five challenging tracking benchmarks, including VideoCube, LaSOT, LaSOT$_{ext}$, TNL2K, and GOT-10k. 
Table~\ref{tab:sota-eval} presents all the tracking results.

\vspace{1mm}
\noindent \textbf{VideoCube}~\cite{videocube}. VideoCube is a comprehensive and challenging long-term visual tracking benchmark designed to reflect the complexities of the real world, such as object occlusion and disappearances.
It comprises 500 video segments, each containing at least 4,008 frames, averaging about 14,920.
This benchmark introduces a novel global instance tracking task, where the tracker should locate a specified instance in a video without assuming consistent camera or motion patterns.
Our tracker is evaluated solely on its test set, comprising 100 sequences.
As shown in Table~\ref{tab:sota-eval}, our approach achieves the best performance compared to the state-of-the-art methods.
Compared to the second-best tracker MixViT~\cite{cui2023mixformer}, our method achieves a 2.8\% increase in normalized precision (NP) and a 2.4\% gain in success rate (SUC).
The positive performance highlights the promising potential of our tracker to effectively address challenges related to camera switching and occlusion, which benefits from the association between the tracker and detector through accurate target state prediction.

\begin{table*}[t]
\footnotesize
\rowcolors{2}{gray!25}{white}
\begin{center}
\caption{\textbf{State-of-the-art comparisons on the datasets of VideoCube, TNL2K, LaSOT, LaSOT$_{ext}$, and GOT-10k.} The best two results are shown in \textcolor[rgb]{1,0,0}{red} and \textcolor[rgb]{0,0,1}{blue} color. Our approach performs favorably against the state-of-the-art methods on all datasets.}
 \vspace{-2mm}
 \label{tab:sota-eval}

\renewcommand\arraystretch{1.1}
\begin{tabular}{ll*{13}{c}}
\toprule 
\multirow{2}*{Method} & \multicolumn{3}{c}{VideoCube~\cite{videocube}} & \multicolumn{2}{c}{LaSOT~\cite{LaSOT}}  & \multicolumn{3}{c}{LaSOT$_{ext}$ ~\cite{LaSOT_ext}} & \multicolumn{2}{c}{TNL2K~\cite{TNL2K}} & \multicolumn{3}{c}{GOT-10k~\cite{GOT10k}}

\\\cmidrule(lr){2-4} \cmidrule(lr){5-6}\cmidrule(lr){7-9}\cmidrule(lr){10-11}\cmidrule(lr){12-14}
   & P &NP & SUC & AUC & NP & AUC  & NP & P  & P & SUC & AO  & SR$_{0.75}$ & SR$_{0.5}$\\\midrule
  SiamFC~\cite{SiameseFC} & 3.1 & 12.9 & 6.1  & 33.6 & 42.0 & 23.0 & 31.1 & 26.9 & 28.6 & 29.5 & 34.8 & 39.8 & 35.3\\

 RPN++~\cite{SiamRPN++} & - & - & -  & 49.6 & 56.9  & 34.0 & 41.6 & 39.6 & 41.2 & 41.3 & 51.7 & 32.5 & 61.6\\

 Ocean~\cite{OCEAN} & 18.3 & 51.9 & 32.5  & 56.0 & 65.1  & - & - & - & 37.7 & 38.4 & 61.1 & 47.3 & 72.1 \\

 TransT~\cite{TransT} & - & - & -   & 64.9 & 73.8  & - & - & - & 51.7 & 50.7 & 67.1 & 60.9 & 76.8\\

 KeepTrack~\cite{KeepTrack} & 35.9 & 68.7 & 50.6  & 67.1 & 77.2 & 48.2 & - & - & - & -  & - & - & - \\

 GlobalTrack~\cite{KeepTrack} & 29.3 & 63.1 & 44.8  & 51.7 & 59.7 & 35.6 & 43.6 & 41.1 & 38.6 & 40.5  & - & - & - \\

 Stark~\cite{Stark} & - & - & - & 67.1 & 77.0 & - & - & - & - & - & 68.8 & 64.1 & 78.1\\

 OSTrack~\cite{OSTrack}& - & - & - & 71.1 & 81.1 & 50.5 & 61.3 & 57.6 & - &55.9 & 73.7 & 70.8 & 83.2 \\


 CiteTracker~\cite{citetracker} & - & - & - & 69.7 & 78.6 & - & - & - & 59.6  & 57.7 & 74.7 & 73.0 & 84.3\\ %

 DropTrack~\cite{droptrack} & - & - & - & 71.8 & 81.8 & 52.7 & 63.9 & 60.2 & 57.9  & 56.9 & 75.9 & 72.0 & 86.8\\ %

 MITS~\cite{mits} & 36.9 & 66.7 & 46.6 & 72.1 & 80.1 & 50.3 & 60.1 & 58.6 & 58.5 & 55.5 & \textcolor[rgb]{1,0,0}{80.4} & 75.9 & \textcolor[rgb]{1,0,0}{89.7}\\

 SwinTrack~\cite{SwinTrack} & - & - & -  & 71.3 & -  & 49.1 & - & 55.6 & 55.7 & 55.6 & 72.4 & 67.8 & 80.5\\\midrule

SeqTrack-L~\cite{SeqTrack} & \textcolor[rgb]{0,0,1}{60.9} & 77.4 & 66.1  & 72.5 & 81.5 & 50.7 & 61.6 & 57.5 & - & 57.8 & 74.8 & 72.2& 81.9 \\

MixViT-L~\cite{cui2023mixformer} & 59.9 & \textcolor[rgb]{0,0,1}{78.7} & \textcolor[rgb]{0,0,1}{67.2}  & 73.3 & \textcolor[rgb]{0,0,1}{82.8} & 50.9 & 61.0 & 57.9 & 61.7 & 59.0 & 75.7 & 75.1 & 85.3\\

ARTrack-L~\cite{artrack} & 31.3 & 53.2 & 39.5 & 73.1 & 82.2 & 52.8 & 62.9 & 59.7 & - & \textcolor[rgb]{0,0,1}{60.3} & \textcolor[rgb]{0,0,1}{78.5} & \textcolor[rgb]{1,0,0}{77.8} & \textcolor[rgb]{0,0,1}{87.4}\\

UNINEXT-H~\cite{UNINEXT} & - & - & - & 72.2 & 80.8 & \textcolor[rgb]{1,0,0}{56.2} & \textcolor[rgb]{0,0,1}{63.8} & \textcolor[rgb]{1,0,0}{63.8} & \textcolor[rgb]{0,0,1}{62.8} & 59.3 & - & - & -\\

\midrule
 RTracker-L & \textcolor[rgb]{1,0,0}{63.2} & \textcolor[rgb]{1,0,0}{81.5} & \textcolor[rgb]{1,0,0}{69.6} & \textcolor[rgb]{1,0,0}{74.7}  & \textcolor[rgb]{1,0,0}{84.5} & \textcolor[rgb]{0,0,1}{54.9} & \textcolor[rgb]{1,0,0}{65.5} & \textcolor[rgb]{0,0,1}{62.7} & \textcolor[rgb]{1,0,0}{63.7} & \textcolor[rgb]{1,0,0}{60.6} & 77.9 & \textcolor[rgb]{0,0,1}{76.9} & 87.0
\\\bottomrule
\end{tabular}
\end{center}
\vspace{-5mm}
\end{table*}


\vspace{1mm}
\noindent \textbf{LaSOT}~\cite{LaSOT}. LaSOT is a high-quality, large-scale benchmark for long-term single-object tracking, featuring an average video length exceeding 2,500 frames.
It offers diverse real-world challenges, including scenarios where target objects can intermittently disappear and reappear in view.
As shown in Table~\ref{tab:sota-eval}, our tracker improves all metrics, \eg. 1.4\% in Area Under the Curve (AUC) compared with MixViT and ARTrack.
The encouraging performance demonstrates that our tracker can tackle the object disappearing and reappearing situation, which shows the self-recovery ability.

\vspace{1mm}
\noindent \textbf{LaSOT$_{ext}$}~\cite{LaSOT_ext}. LaSOT$_{ext}$ extends LaSOT with 150 additional videos, introducing the tracking challenges due to similar distractors.
UNINEXT employs a more robust backbone, ViT-Huge, and diverse training datasets for feature extraction, achieving a promising 56.2\% AUC score. 
In comparison, our approach only uses the ViT-Large as the backbone to save computation costs while delivering a comparable performance with a 54.9\% AUC score.
This benefits from the PN tree memory, adeptly storing target-relevant and irrelevant samples like similar distractors in the background, to reliably ascertain the target states for associating the tracker and detector.


\vspace{1mm}
\noindent \textbf{TNL2K}~\cite{TNL2K}. TNL2K is a challenging tracking benchmark that employs template patches and language descriptions to locate target objects in video sequences, facilitating the connection between local and global searches.
However, we only use the bounding box for evaluation.
Compared to UNINEXT with a more robust backbone, our approach still demonstrates improvements with a 0.9\% increase in precision and a 1.3\% increase in SUC, attributed to the RTracking framework that facilitates the connection between local and global search.

\vspace{1mm}
\noindent \textbf{GOT-10k}~\cite{GOT10k}. GOT-10k is a large-scale tracking benchmark covering most categories for over 560 real-world moving objects.
The ground truths for the test set are withheld, and we assess our approach using the evaluation platform provided by the authors.
We follow the one-shot rule with zero overlapping in object classes between the training and test sets and use the tracker and the detector trained only on the GOT-10k training set.
Our  RTracker also achieves competitive results, comparable to the most recent trackers MITS~\cite{mits} and ARTrack~\cite{artrack}.
The good performance shows that our tracker has a good generalization ability to the tracking scenarios involving class-agnostic targets.

\begin{table}[]
\begin{center}   
 \caption{\textbf{Ablation study of the proposed algorithm on the VideoCube, LaSOT$_{ext}$, and TNL2K datasets. }
 The best results are marked in \textbf{bold}.}
 \vspace{-2mm}
 \rowcolors{2}{gray!25}{white}
  \label{tab:ablation}
 \footnotesize

\begin{tabular}{ccccccc}
\midrule
 \multirow{2}*{Method} & \multicolumn{2}{c}{VideoCube} & \multicolumn{2}{c}{LaSOT\_ext}  & \multicolumn{2}{c}{TNL2K}

\\\cmidrule(lr){2-3} \cmidrule(lr){4-5}\cmidrule(lr){6-7}

 & AUC  & NP  & AUC  & NP & AUC & NP \\ \midrule
Base T   & 67.2  & 78.7 & 50.7 & 61.6 & 59.0  & 75.5 \\
Base D   & 46.6  & 66.7 & 50.3 & 60.1 & 55.5  & 69.7 \\
Fixed THR   &60.9 & 72.1 & 50.7 & 59.9   & 56.5  & 71.5  \\
W/O WR   & 62.6 & 73.6  & 49.0  & 58.0 & 57.4 & 73.3  \\ \midrule
RTracker & \textbf{69.6} & \multicolumn{1}{l}{\textbf{81.5}} & \multicolumn{1}{l}{\textbf{54.9}} & \multicolumn{1}{l}{\textbf{65.5}} & \multicolumn{1}{l}{\textbf{60.6}} & \multicolumn{1}{l}{\textbf{76.9}} \\ \midrule
\end{tabular}
\end{center}
\vspace{-6mm}
\end{table}


 

\subsection{Ablation Study}
To evaluate the effect of each individual component of our tracker, we carry out ablation studies on five different variants of the RTracker:

\vspace{1mm}
\noindent \textbf{RTracker,} our intact model dynamically associates a tracker and a detector that offers global search capabilities to achieve recoverable tracking based on the changing target states predicted by relative measurements utilizing the PN tree memory.

\vspace{1mm}
\noindent \textbf{Base Tracker (T),} which employs only the tracker in RTracker to track targets.
In this work, we use the MixViT-L~\cite{cui2023mixformer} as our base tracker.

\vspace{1mm}
\noindent \textbf{Base Detector (D),} which employs only the detector in RTracker to track targets.
In this work, we use the MITS~\cite{mits} as our base detector.

\vspace{1mm}
\noindent \textbf{Fixed Threshold (THR),} which predicts the current state of the target by comparing the tracking confidence score with a pre-set fixed threshold in place of the state prediction by the PN tree.

\vspace{1mm}
\noindent \textbf{W/O Walking Rules (WR),} which determines the current state of the target based on the relative distances to temporally adjacent positive and negative samples instead of the state prediction through the walking rules.

Table~\ref{tab:ablation} presents the experimental results of these variants on the VideoCube~\cite{videocube}, LaSOT$_{ext}$~\cite{LaSOT_ext}, and TNL2K~\cite{TNL2K} datasets.

\vspace{1mm}
\noindent \textbf{Effect of associating tracking and detection.}  
By comparing our RTrack with the base tracker, it is clear that the proposed tracking algorithm improves tracking performance by $2.8\%$, $3.9\%$, and $1.4\%$ in terms of NP on VideoCube, LaSOT$_{ext}$ and TNL2K, respectively. 
While compared with the base detector, our tracker achieves performance gain of $23\%$, $4.6\%$, and $5.1\%$ in terms of AUC on VideoCube, LaSOT$_{ext}$ and TNL2K, respectively.
The gap performance between the base tracker, the base detector, and our proposed method demonstrates the effectiveness of our approach in associating the tracker and detector.

\begin{figure}[t]
\begin{center}
\includegraphics[width=1.0\linewidth]{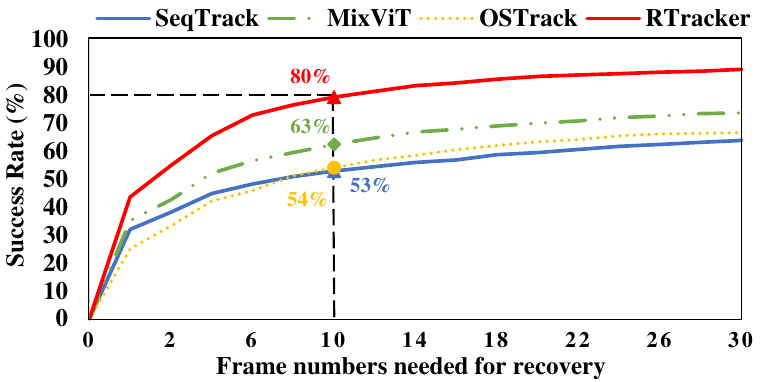}
 \caption{\textbf{Evaluation of the recovery ability on LaSOT.}
 The success rate is the percentage at which a tracker successfully recovers disappeared targets within specific frame numbers. 
 }
 \vspace{-2mm}
\label{Fig:recovery}
\end{center}
\vspace{-5mm}
\end{figure}

\vspace{1mm}
\noindent \textbf{Effect of the PN tree.}
With the relative measurement between the positive and negative features stored in the PN tree, RTrack achieves performance gains of $8.7\%$, $4.2\%$, and $4.1\%$ in AUC on VideoCube, LaSOT$_{ext}$, and TNL2K, while $9.4\%$, $5.6\%$ and $5.4\%$ in NP, respectively. 
The enhancements observed validate the benefits of relative measurements.
Unlike a fixed threshold for ascertaining the target state, relative measurements provide favorable adaptability to target changes during tracking, enhancing the robustness of target state predictions.

\vspace{1mm}
\noindent \textbf{Effect of using the walking rules.} Without the walking rules, RTracker decrease by $7\%$, $5.9\%$ and $3.2\%$ in precision on VideoCube, LaSOT$_{ext}$ and TNL2K. 
The results confirm the effectiveness of our method in employing walking rules to continually update memory, capture appearance changes, and accurately determine target states.
%

\begin{figure*}[!t]
\centering
\includegraphics[width=0.98\linewidth]{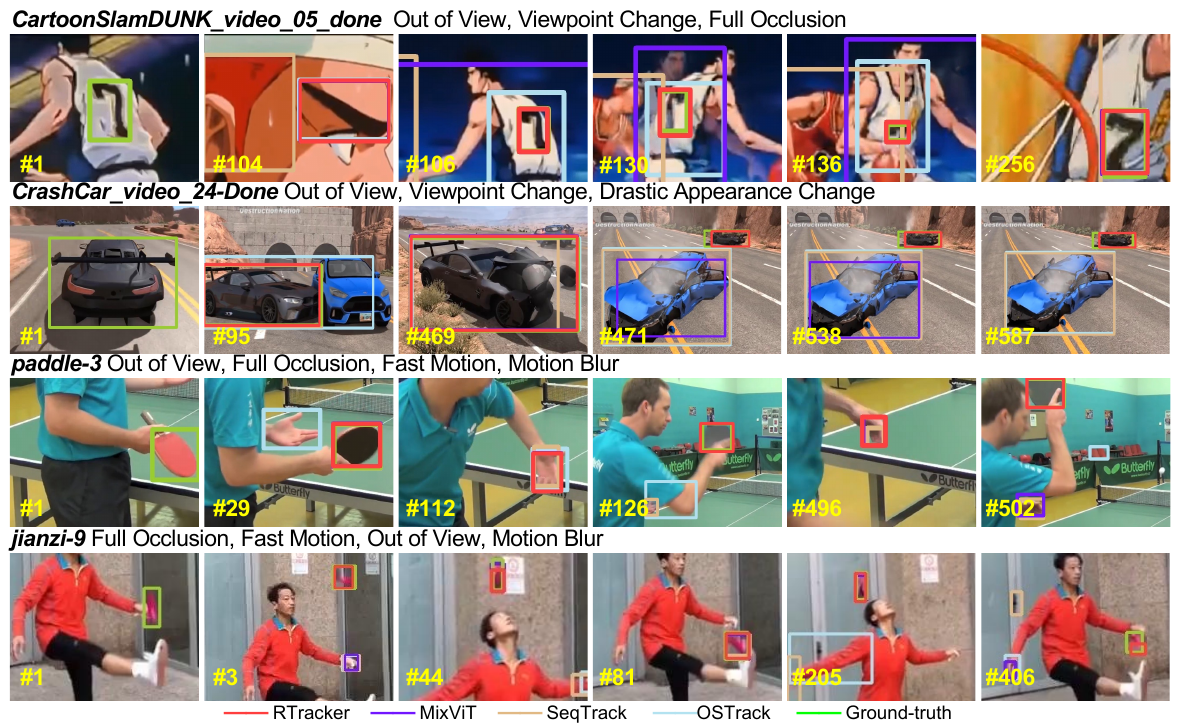}
\vspace{-4mm}
 \caption{\textbf{Visualized results of the proposed algorithm, MixViT, OSTrack, and SeqTrack method on four challenging sequences with drastic changes.}
 {This indicates that our RTracker performs well with the support of detection through PN tree memory, whereas the other method that relies only on the tracker faces difficulties with these sequences.}
 }
\label{Fig:case}
\vspace{-5mm}
\end{figure*}

\subsection{Recovery Ability Evaluation}
We assess the recovery ability of the proposed method on the LaSOT dataset by measuring the number of frames required to relocate a lost target.
A recovery is deemed successful if the overlap between the predicted bounding box and its ground truth exceeds 0.5.
As illustrated in Figure~\ref{Fig:recovery}, RTracker can recover more lost targets within the same time as other trackers.
Notably, RTracker achieves a high success rate, successfully recovering targets in 80\% of the test cases where they were lost.
In comparison, MixViT has a 63\% success rate in recovering targets, while SeqTrack and OSTrack exhibit average performance, managing successful recovery in about half of the target loss cases. 
The comparison shows that RTracker, combining tracking and detection with PN tree memory, has enhanced target recovery ability in challenging situations and is more effective than other methods reliant solely on tracking.

\subsection{Qualitative Study}
To demonstrate the self-recovery capability of our proposed method, we visualize the tracking results of several challenging sequences with MixViT, SeqTrack, and OSTrack.
In the $\it{CartoonSlamDUNK}$ sequence, RTracker can immediately relocate to the position of the target after it gets lost at the 104\textit{th} frame and recovers at the 106\textit{th} frame, whereas other trackers can only relocate the target after the 130\textit{th} frame.
In the $\it{CrashCar}$ series, our tracker accurately tracks the target despite shifts in the tracking viewpoint, whereas other trackers may not.
The visualized results demonstrate that RTracker can recover from the missing or out-of-view targets due to the association of the tracker and detector via the target states. 
Moreover, our RTracker can accurately locate the target even if the target undergoes drastic appearance changes or is blurred.
In the $\it{paddle}$ sequence, the color of the target changes from red to black by the 29\textit{th} frame. Our tracker still tracks the target correctly, whereas other trackers might locate the area more similar to the first frame.
Despite the fast motion and full occlusion in the $\it{jianzi}$ sequence, our tracking method still performs well.

\begin{figure}[t]
\begin{center}
\includegraphics[width=0.98\linewidth]{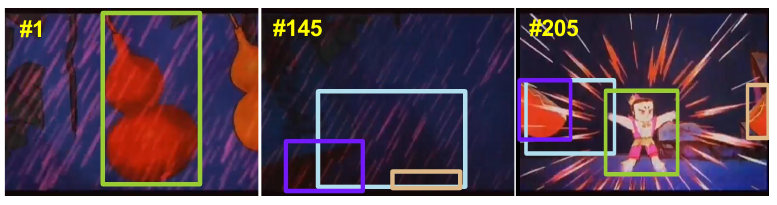}
\vspace{-2mm}
 \caption{\textbf{Failure case of the proposed method. } 
The target disappears at frame 145 and recovers in a completely different appearance at frame 205, with similar distractors in the background.
 }
\label{fig:limitation}
\end{center}
\vspace{-5mm}
\end{figure}

\subsection{Limitations}
%
Due to the additional computational loads for evaluating the present state of the target and facilitating integration with the detector, RTracker runs at a speed of 0.75 times that of the baseline method.
We intend to reduce the effects of this extra cost by using lightweight techniques. 
In addition, RTracker may not be able to recover tracking correctly and timely in a few extreme scenarios where the target reappears with a totally different appearance, and there are also similar background distractors simultaneously. 
Figure~\ref{fig:limitation} shows an example where the target disappears at frame 145 and reappears at frame 205, where all trackers fail to track accurately.
To address this issue, our future work will incorporate semantic descriptors to guide the tracker in effectively adapting to targets with entirely different appearances.

\section{Conclusion} 
In this work, we present a recoverable tracking framework, \ourmethod, that dynamically associates a tracker and a detector for self-recovery. 
Specifically, we construct a Positive-Negative Tree Structured memory, which maintains the samples relevant and irrelevant to the tracking target chronologically.
In addition, we formulate a set of walking rules for the PN tree memory, enabling the reliable determination of the target state by assessing the relative distances between positive and negative samples.
Upon this target state, we define three control flows to associate the tracker and the detector adaptively for robust tracking.
Both qualitative and quantitative assessments demonstrate that our approach performs favorably against state-of-the-art methods, highlighting the effectiveness of dynamically integrating trackers and detectors with PN tree memory for improving tracking performance.

\section{Acknowledgments}
\label{sec:Acknowledgments}
The paper is supported by the Major Key Project of PCL (PCL2023A08), the National Natural Science Foundation of China (62172126, 62002241, U20B2052), and the Shenzhen Research Council (No. JCYJ20210324120202006).

{
    \small
    \bibliographystyle{ieeenat_fullname}
    \bibliography{tracking}
}

\input{supp}
\end{document}

%% file: preamble.tex
\usepackage{epstopdf}

%% file: supp.tex
\clearpage
\setcounter{page}{1}
\maketitlesupplementary

\appendix

\renewcommand{\thefootnote}{}
\footnotetext{$\ast$ corresponding author}
This document provides additional information on the experimental implementation and results.

\vspace{-1mm}
\section{Implementation Details}
\label{sec:rationale}

\subsection{Feature Extractor in the PN tree}
We use the features extracted by a similarity perception model~\cite{dreamsim} to describe the state of the tracking targets as they are more robust to changes in the target.

\vspace{1mm}
\noindent\textbf{Datasets.}
To adapt the similarity perception model for tracking tasks, we train it on two datasets: Novel Image Generations with Human-Tested Similarity (NIGHTS)~\cite{dreamsim} dataset and LaSOT~\cite{LaSOT} training set. 
NIGHTS comprises human similarity judgments for image triplets, each consisting of a reference image and two altered versions, along with human assessments about which version is most similar to the reference.
For LaSOT, we employ the target found in the initial frame of each sequence as the template. 
Subsequently, we crop the ground truth area of the tracking target from the following frames as A, ensuring that it pertains to the same object as the template.
We also extract targets from other sequences of the same category as B, which is similar to the template.
In total, we exploit 20,000 triplets from NIGHTS and generate an additional 4,000 triplets from LaSOT as our training datasets.

\vspace{1mm}
\noindent\textbf{Training Settings.}
We denote the distance between two samples as $D$ computed as follows:
\begin{align}
D_1 &= 1-cos(f_\theta(Template),f_\theta(A)),\\
D_2 &= 1-cos(f_\theta(Template),f_\theta(B)),
\end{align}
where $f_\theta$ represents the feature extractor, $D_1$ is the distance between the template and A samples while $D_2$ is the distance between the template and B samples.
To improve target state assessment, especially in scenarios with similar noise to tracking targets, we aim to maximize the difference between the distances, $D_1$ and $D_2$, while simultaneously minimizing $D_1$ for the tracking target from the same sequence within the triplet $(Template, A, B)$. 
Therefore, we follow the training settings of dreamsim~\cite{dreamsim} and use a hinge training loss can be formulated as:
\begin{gather*}
\mathcal{L} = max(0, m - \Delta{D}\cdot \bar{y} ), \Delta{D} = D_0 - D_1, \\
\bar{y} =\left\{
\begin{aligned}
1 =& D_1 < D_0\\
0 =& D_0 < D_1 ,
\end{aligned}
\right.
\end{gather*}
where $\bar{y}$ is the relative distance judgment between the template and A/B and m equals to 0.05.
\begin{figure}[t]
\begin{center}
\includegraphics[width=0.7\linewidth]{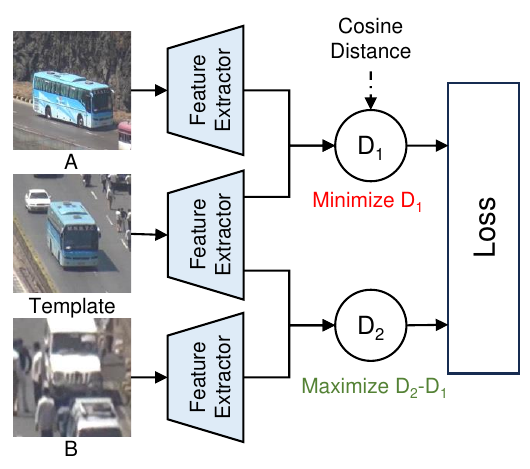}
\vspace{-2mm}
 \caption{\textbf{Training method.} 
Given a triplet, we have a template, A and B samples. 
we compute the cosine distances, $D_1$ and $D_2$, between the features of the Template and A/B. 
Our training objective is to minimize $D_1$ and maximize the gap between $D_1$ and $D_2$.
 }
\label{fig:limitation}
\end{center}
\vspace{-5mm}
\end{figure}

\section{More Detailed Experimental Results} 
\subsection{Detailed results on VideoCube}
Table~\ref{tab:sota-eval} shows the evaluation results on the VideoCube benchmark.
We conduct tests on the VideoCube dataset using the provided checkpoints from SeqTrack-Large~\cite{SeqTrack}, MixViT-Large~\cite{cui2023mixformer}, and MITS~\cite{mits}. 
However, for ARTrack~\cite{artrack}, since the authors did not release the checkpoint of ARTrack-Large,  we independently train ARTrack-Large on our own machine to conduct the testing.
%
%
\vspace{-2mm}
\subsection{Detailed results on LaSOT}
\begin{figure*}[t]
 \begin{minipage}{0.48 \linewidth}
 	\vspace{5pt}
 	\centerline{\includegraphics[width=\textwidth]{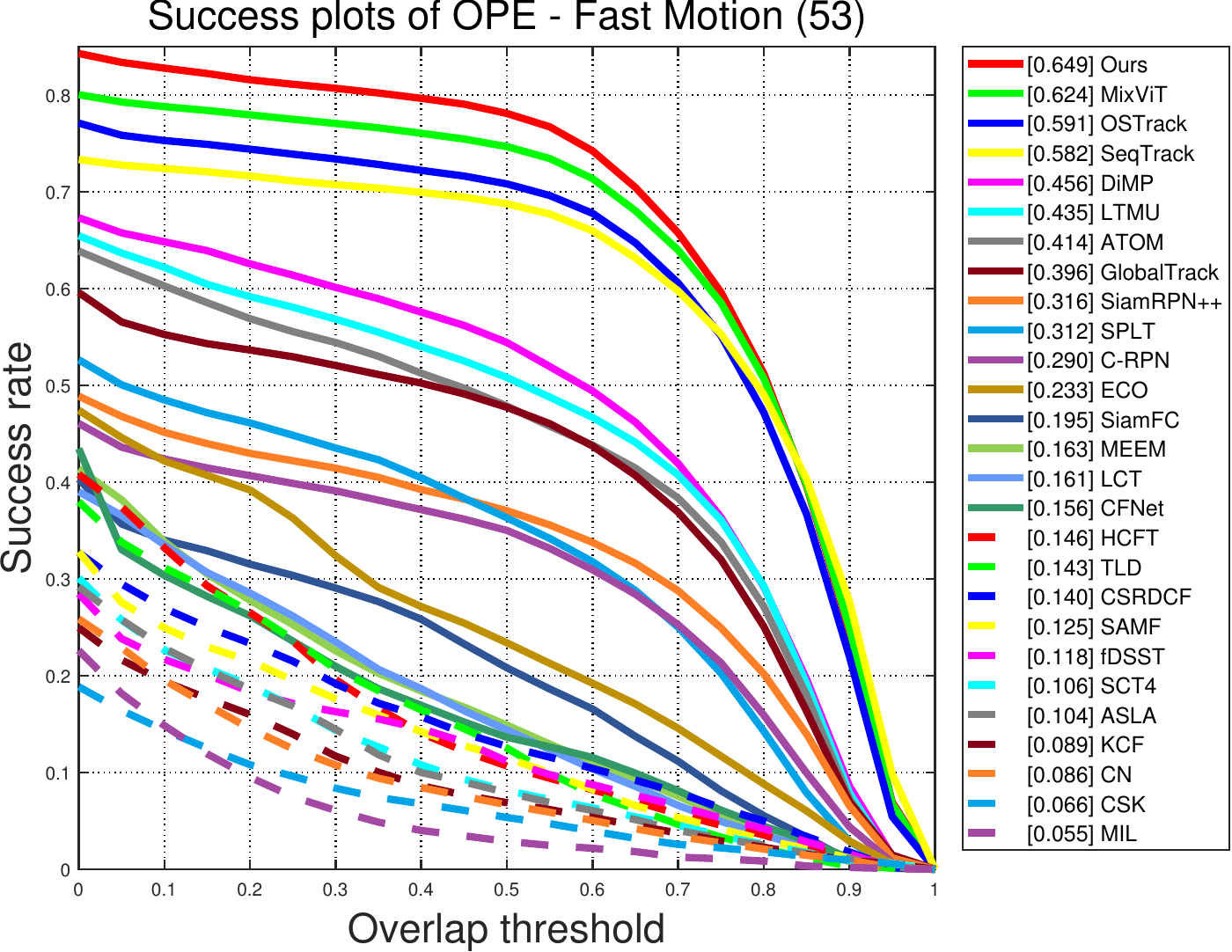}}
  	\vspace{5pt}
	  \centerline{\includegraphics[width=\textwidth]{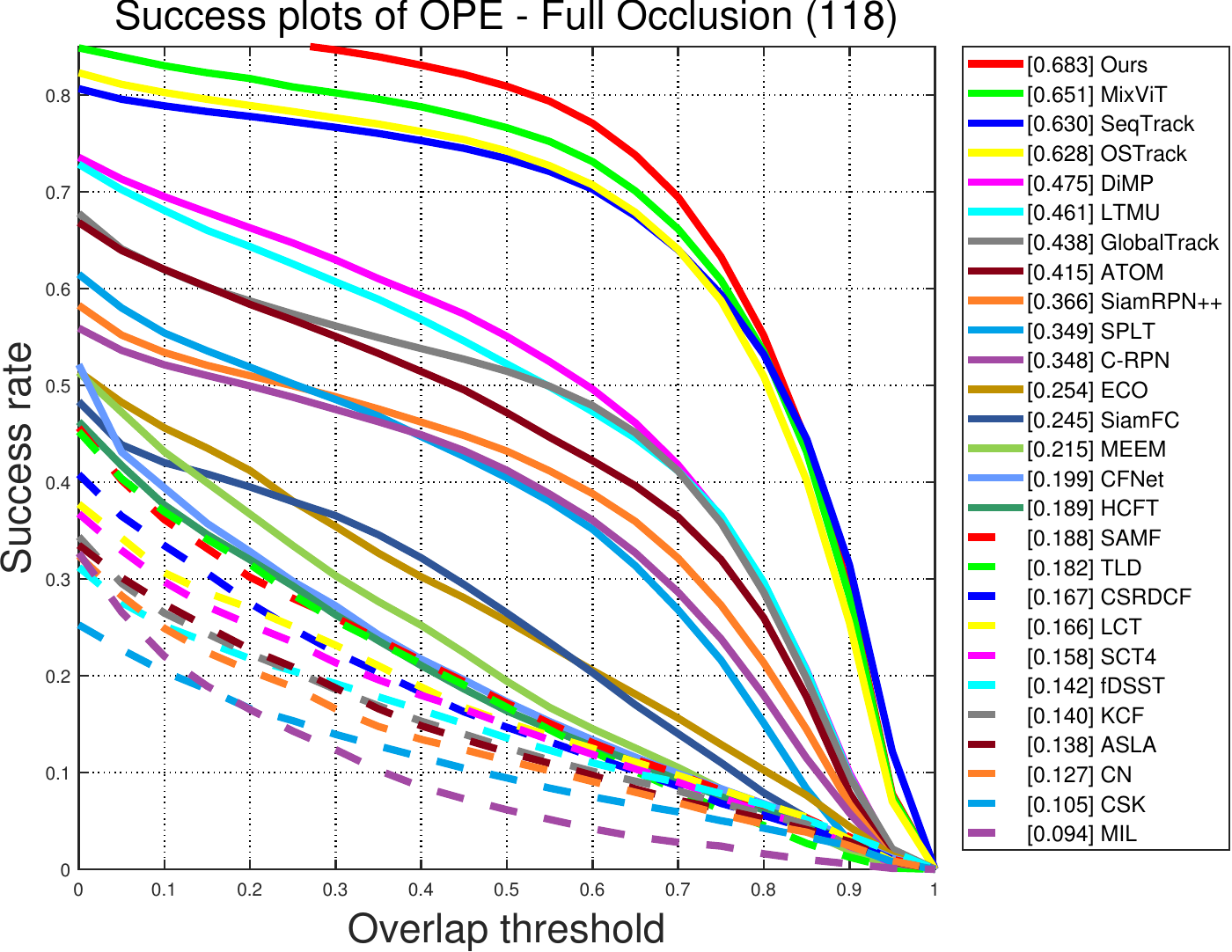}}
	  \vspace{5pt}
	  \centerline{\includegraphics[width=\textwidth]{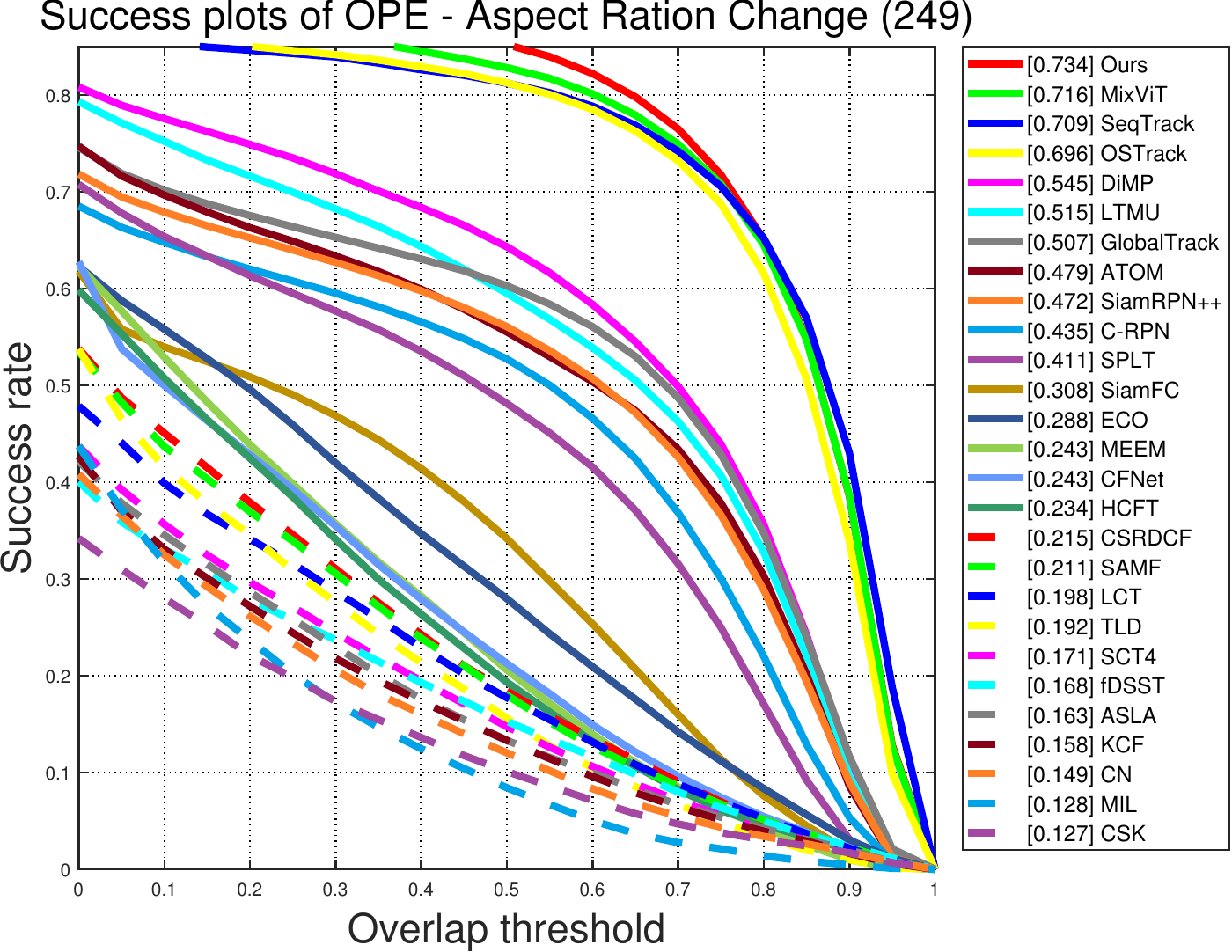}}
 \end{minipage}
 \begin{minipage}{0.48\linewidth}
	\vspace{5pt}
	\centerline{\includegraphics[width=\textwidth]{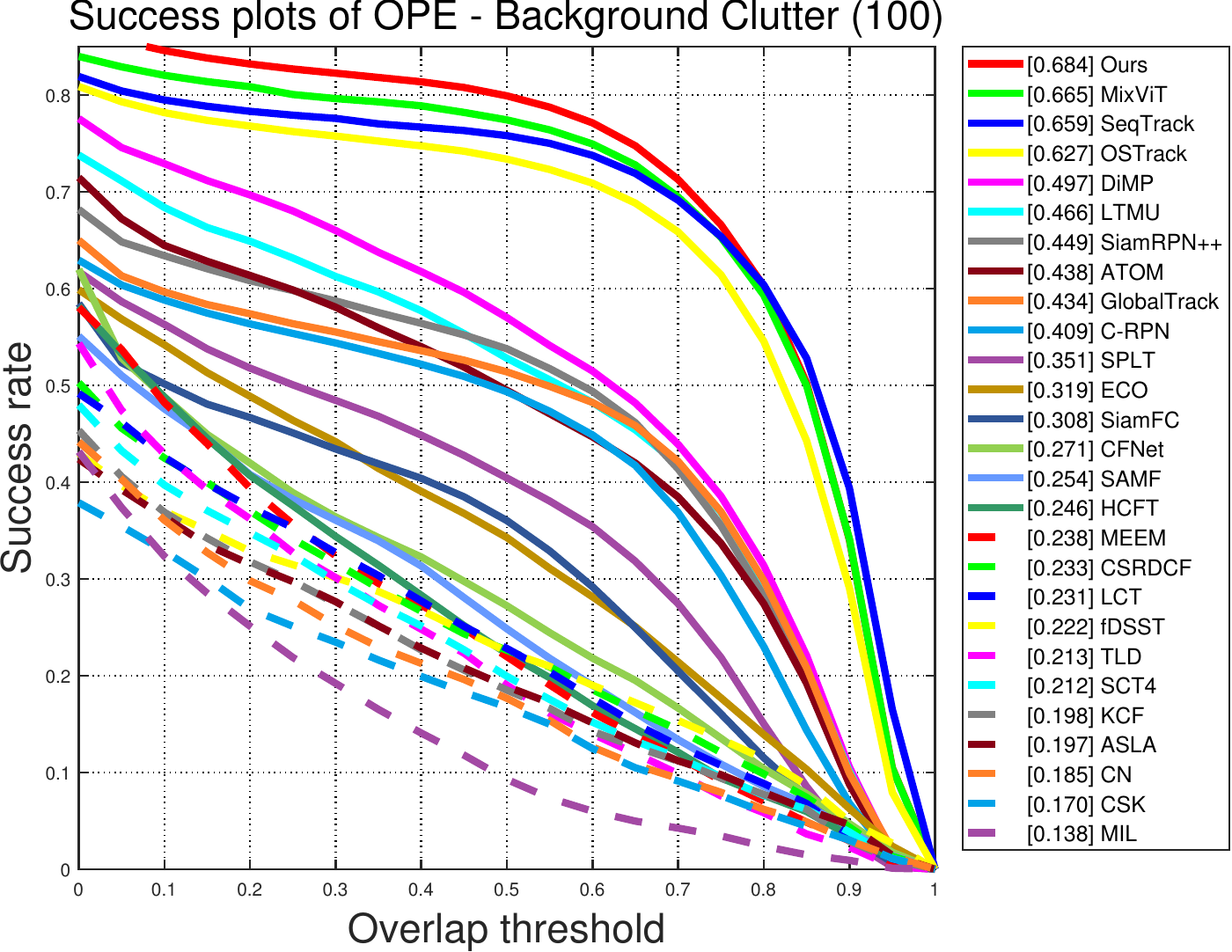}}
    \vspace{5pt}
	\centerline{\includegraphics[width=\textwidth]{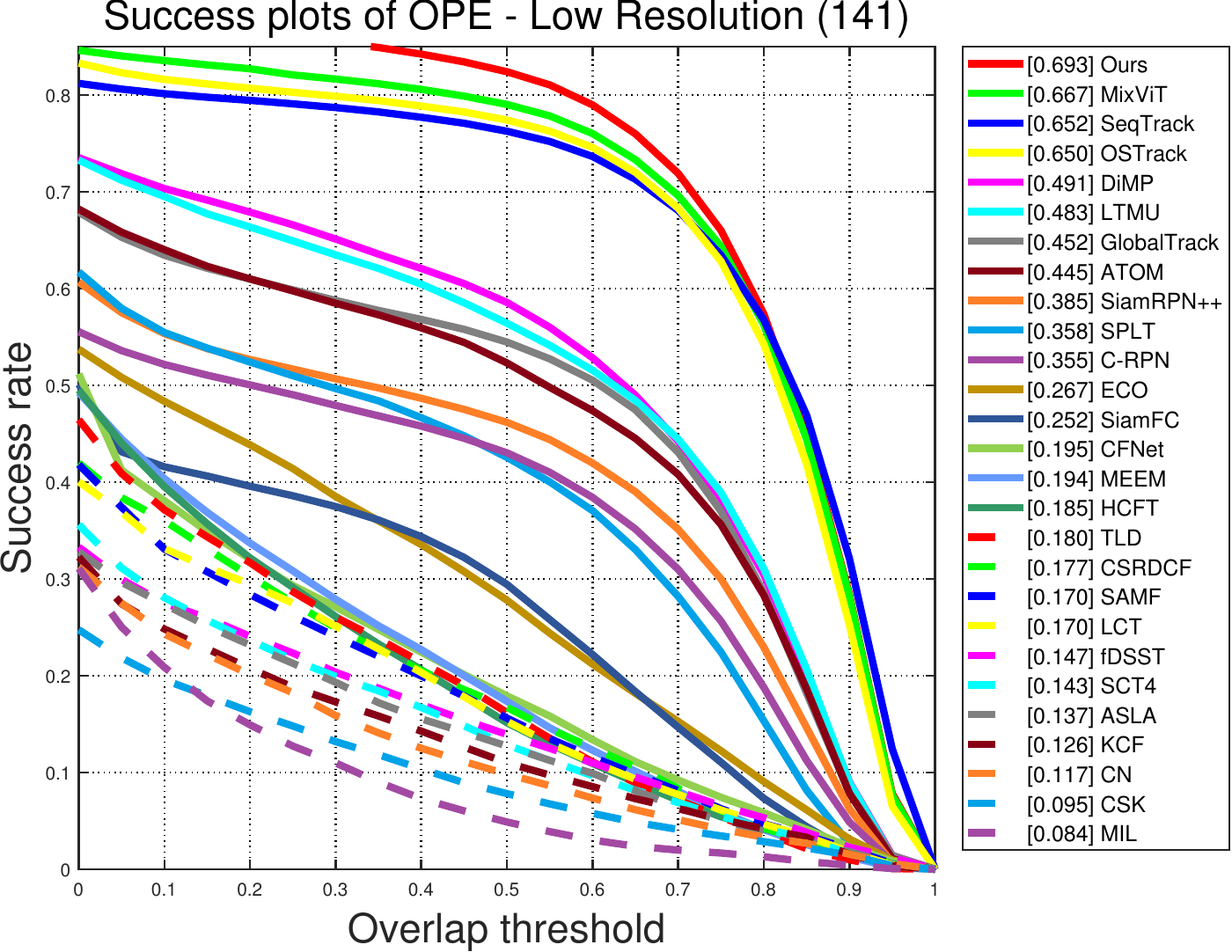}}
	\vspace{5pt}
	\centerline{\includegraphics[width=\textwidth]{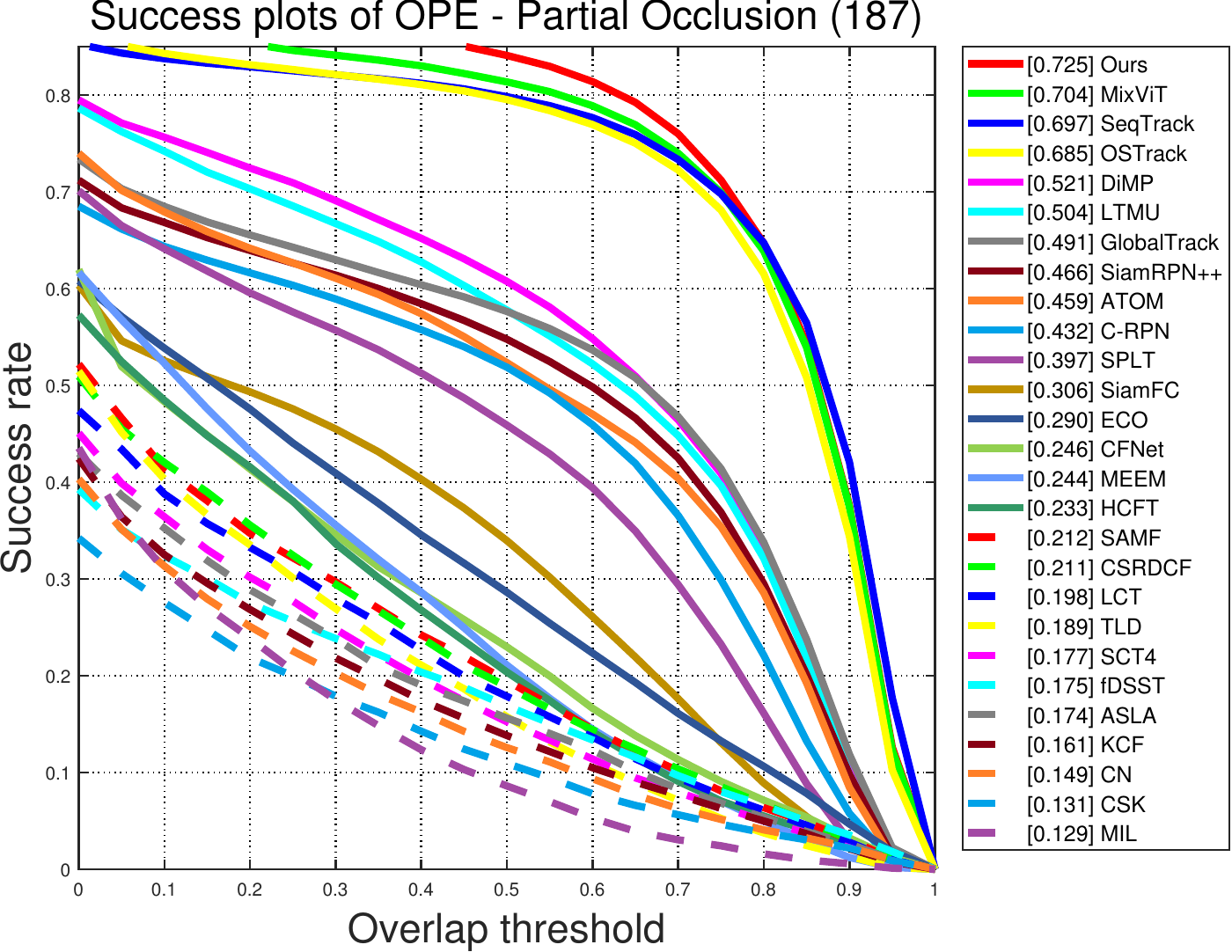}}
\end{minipage}
\caption{\textbf{Success plots of different attributes on LaSOT.} 
}
\label{fig:lasot-1}
\end{figure*}

\begin{figure*}[!t]
 \begin{minipage}{0.48 \linewidth}
 	\vspace{5pt}
 	\centerline{\includegraphics[width=\textwidth]{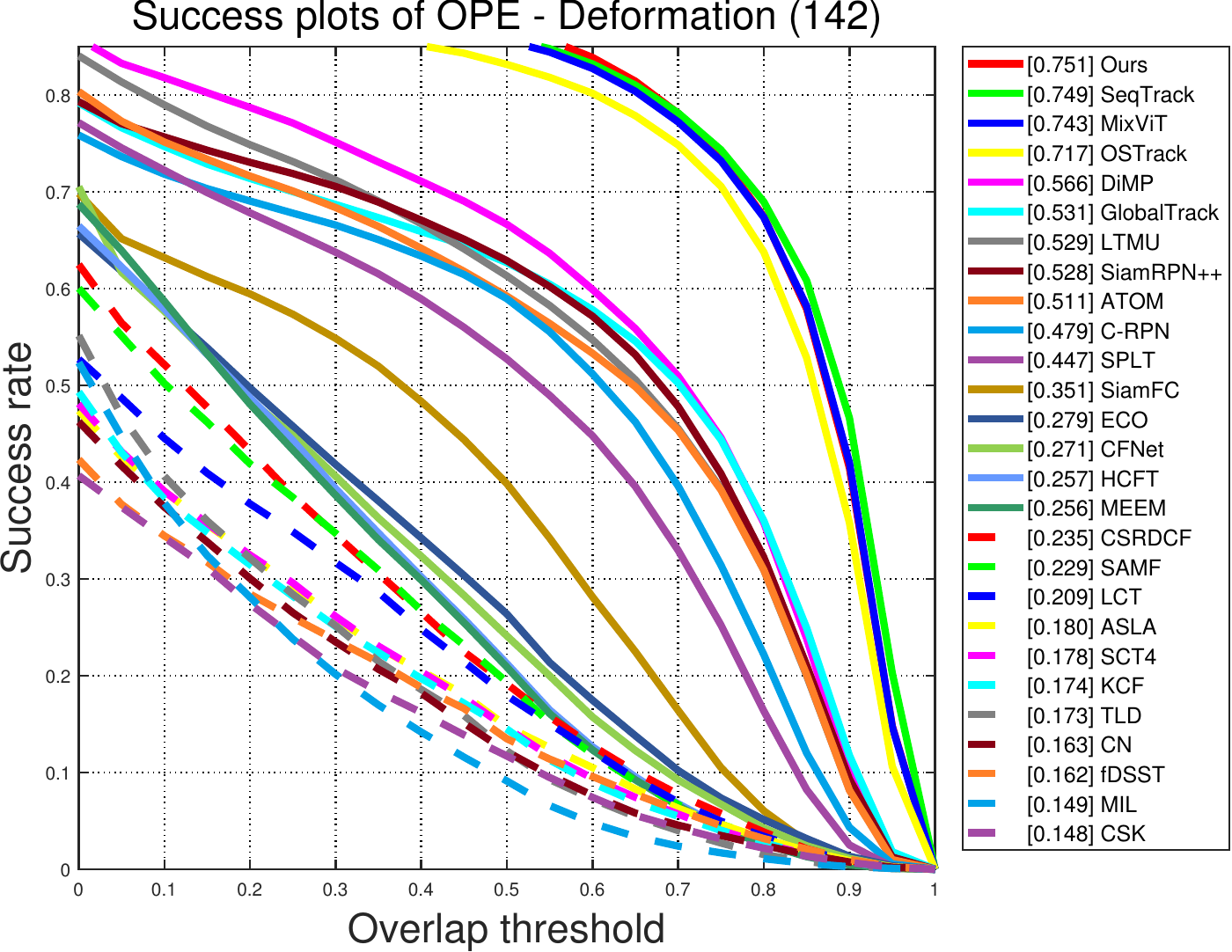}}
  	\vspace{5pt}
	  \centerline{\includegraphics[width=\textwidth]{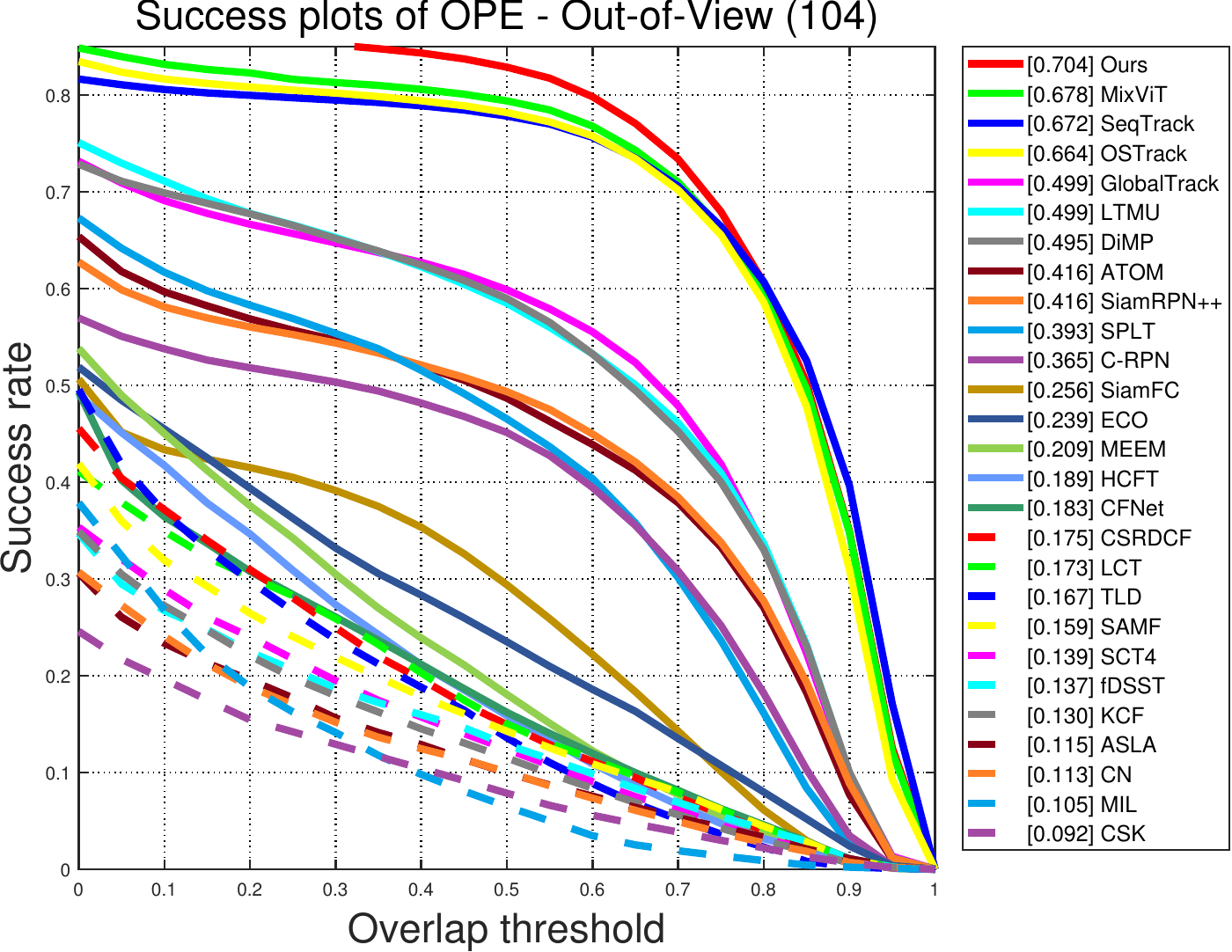}}
	  \vspace{5pt}
	  \centerline{\includegraphics[width=\textwidth]{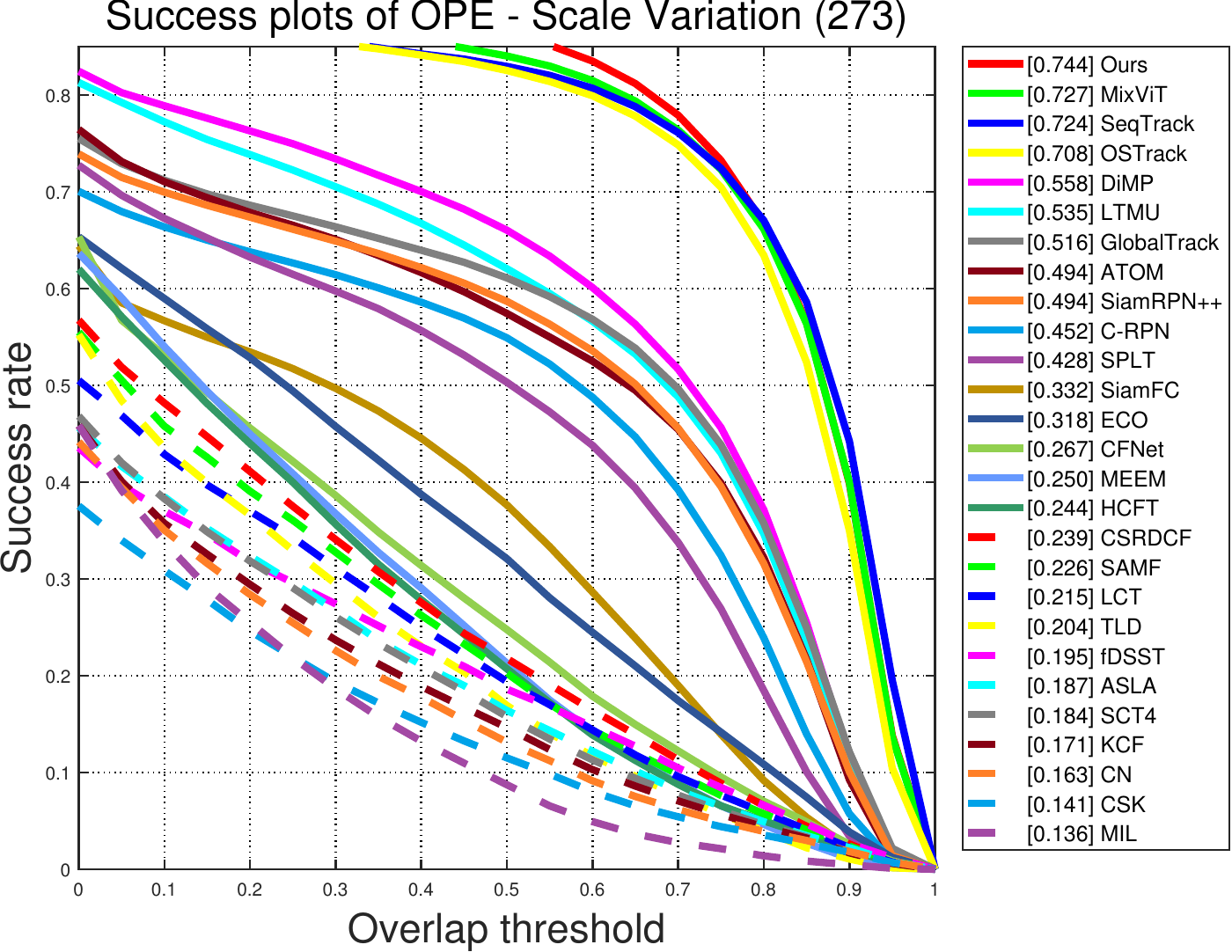}}
 \end{minipage}
 \begin{minipage}{0.48\linewidth}
	\vspace{5pt}
	\centerline{\includegraphics[width=\textwidth]{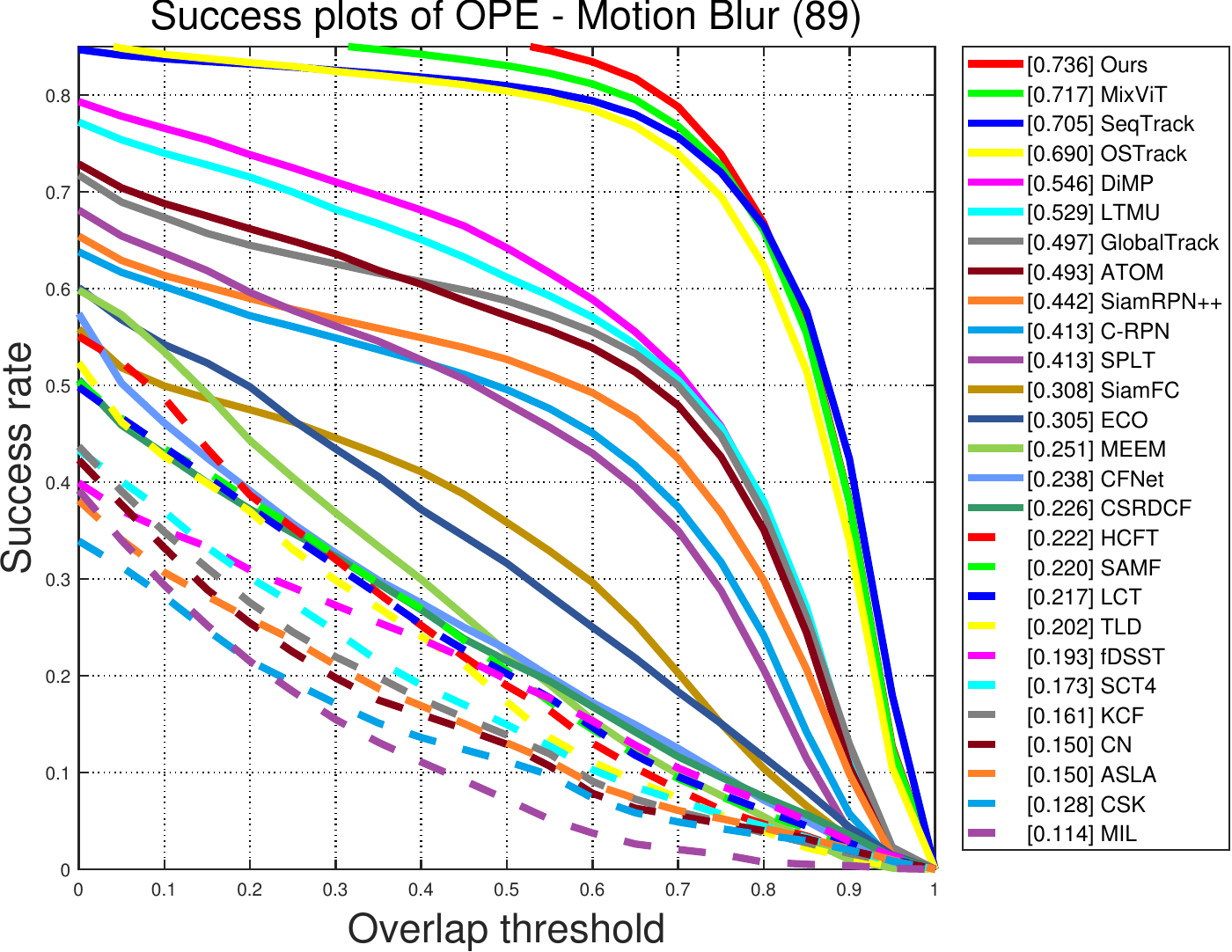}}
    \vspace{5pt}
	\centerline{\includegraphics[width=\textwidth]{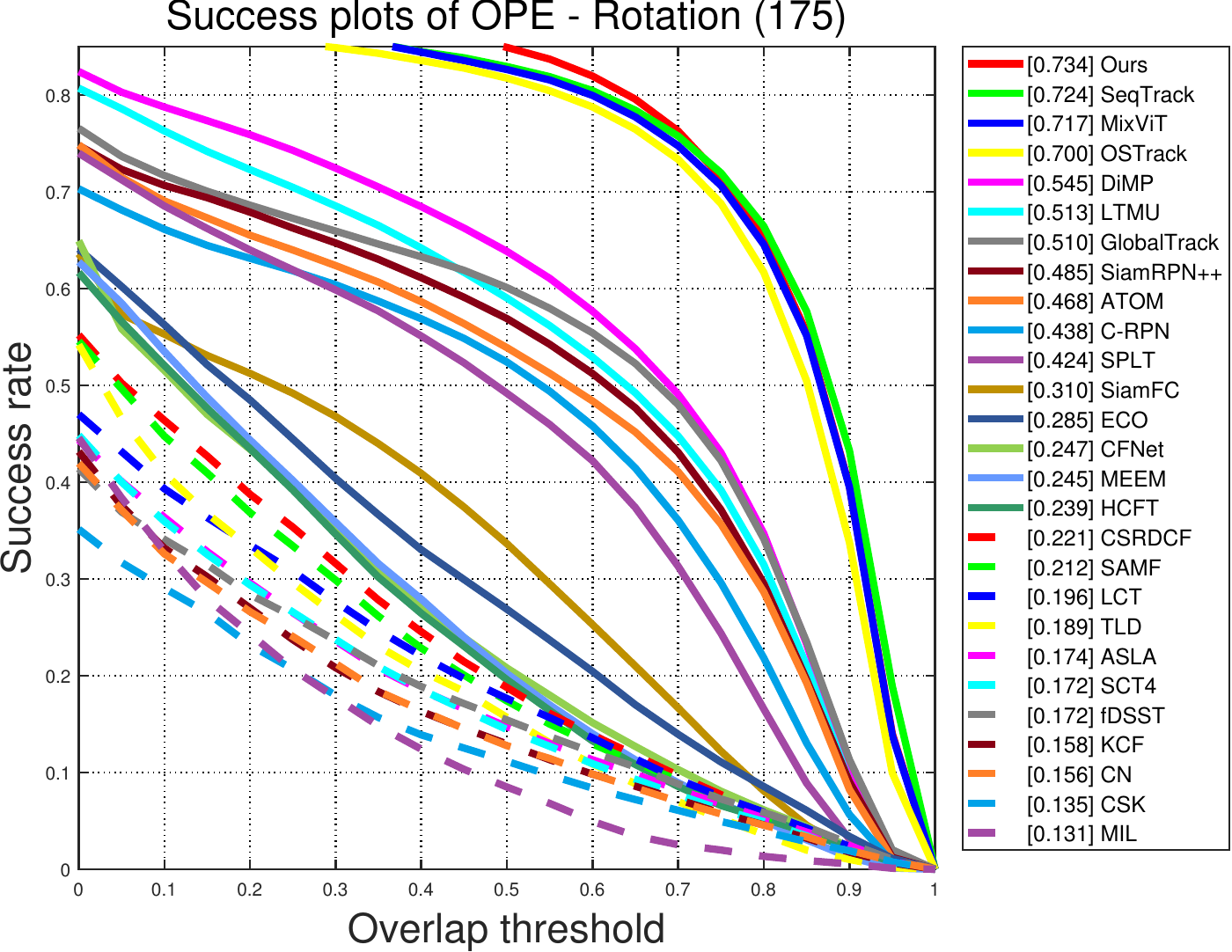}}
	\vspace{5pt}
	\centerline{\includegraphics[width=\textwidth]{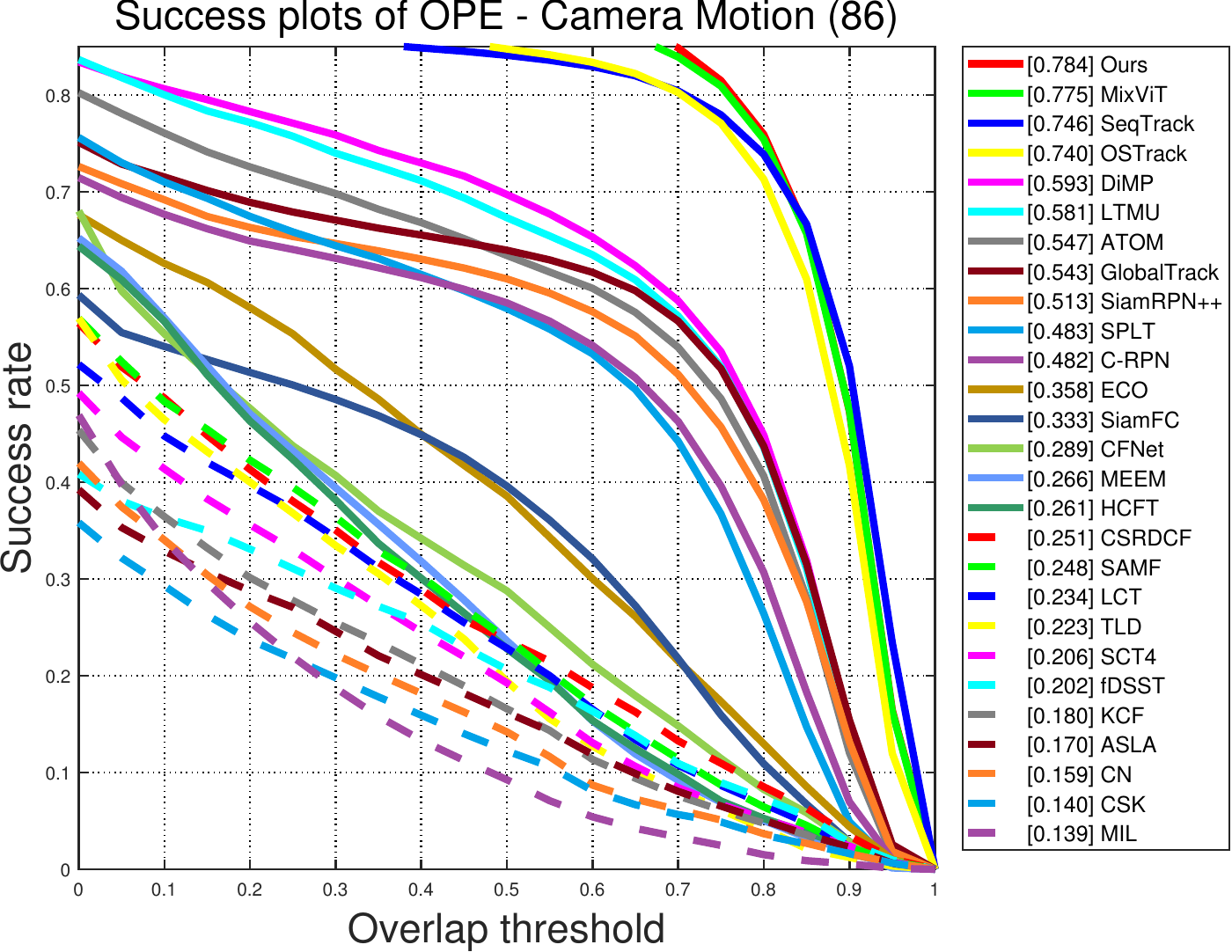}}
\end{minipage}
\caption{\textbf{Success plots of different attributes on LaSOT.}
}
\label{fig:lasot-2}
\end{figure*}

The AUC score under different attributes of LaSOT~\cite{LaSOT} test set is shown in Figure~\ref{fig:lasot-1} and Figure~\ref{fig:lasot-2}. 
Compared to the second-best tracker MixViT, our proposed method achieves an improvement of 3.2\% in out-of-view cases and 2.6\% AUC scores in full-occlusion cases on the LaSOT test set.
This demonstrates that our tracker possesses a robust self-recovery capability, attributed to our approach of combining detection with tracking based on the target states, enabling the tracker to handle situations where the target is missing and swiftly relocate the target.
\vspace{-2mm}
\subsection{Detailed results on LaSOT$_{ext}$}
\begin{figure*}[!t]
 \begin{minipage}{0.48 \linewidth}
 	\vspace{5pt}
 	\centerline{\includegraphics[width=\textwidth]{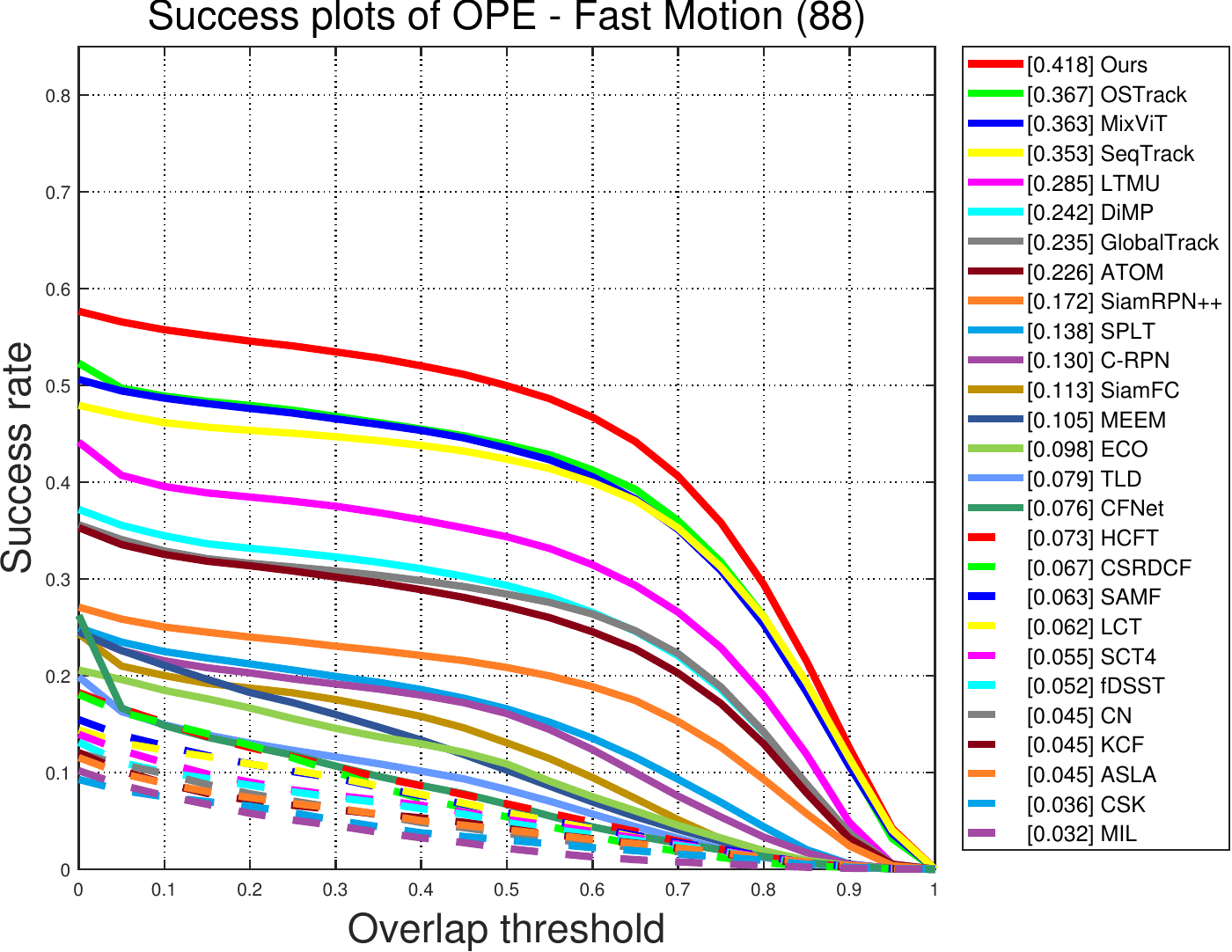}}
  	\vspace{5pt}
	  \centerline{\includegraphics[width=\textwidth]{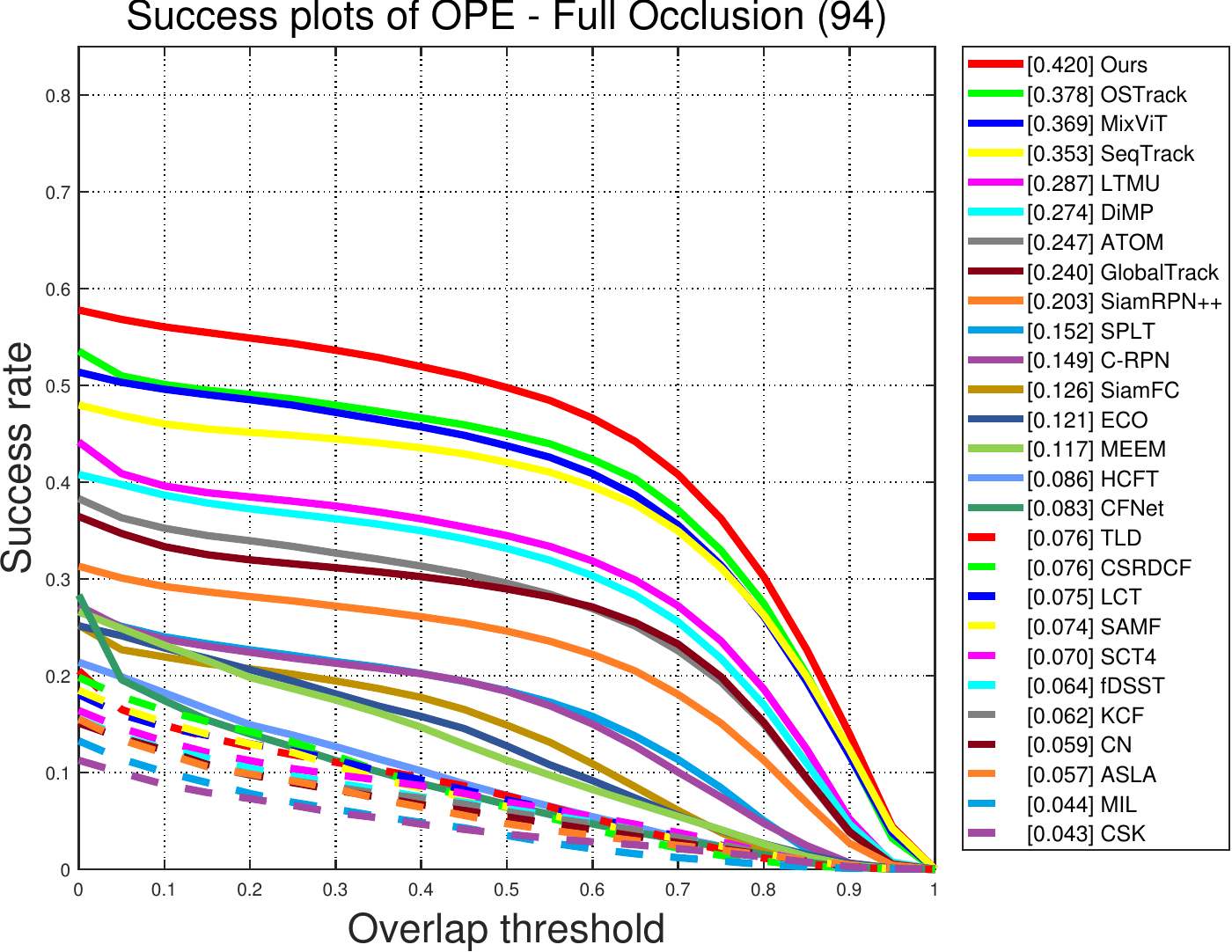}}
	  \vspace{5pt}
	  \centerline{\includegraphics[width=\textwidth]{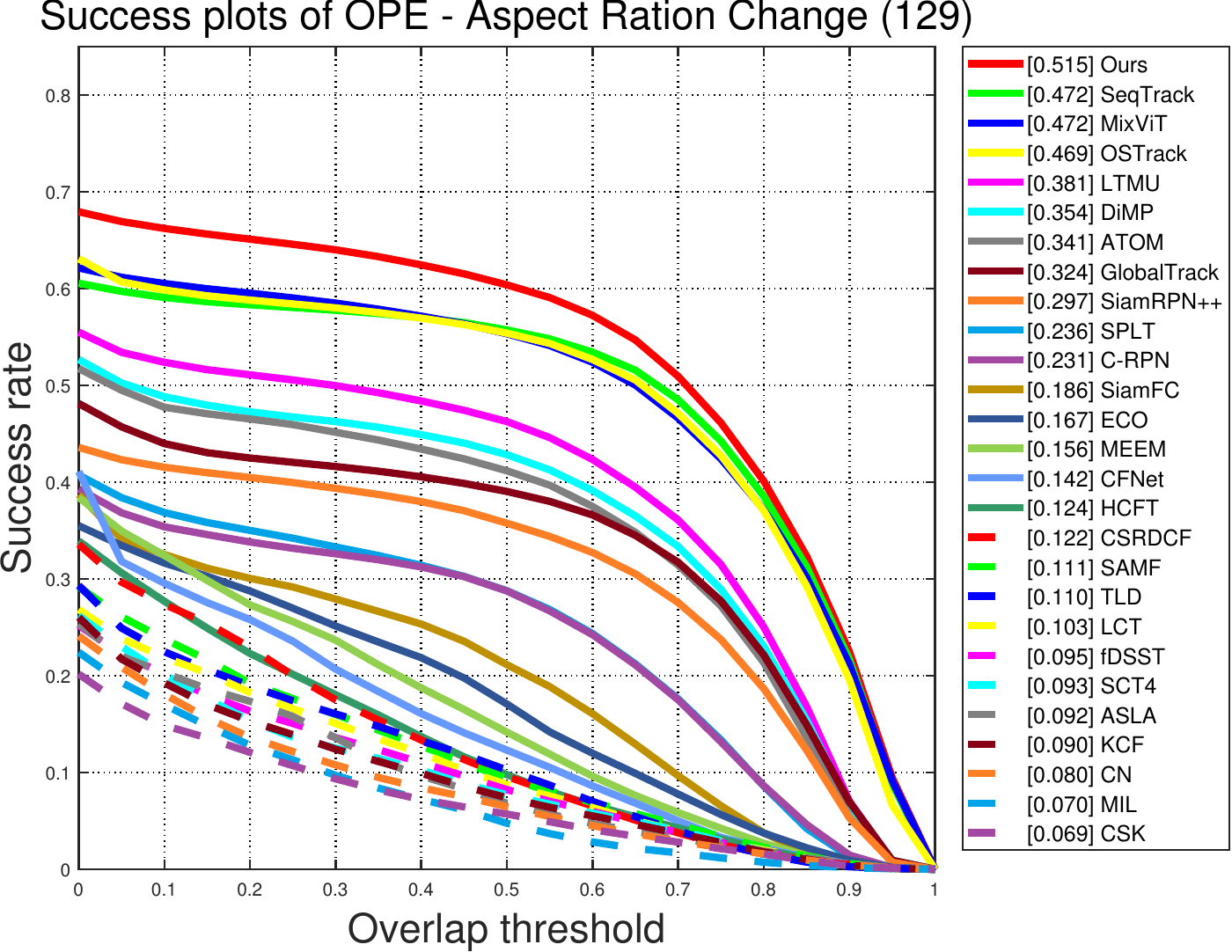}}
 \end{minipage}
 \begin{minipage}{0.48\linewidth}
	\vspace{5pt}
	\centerline{\includegraphics[width=\textwidth]{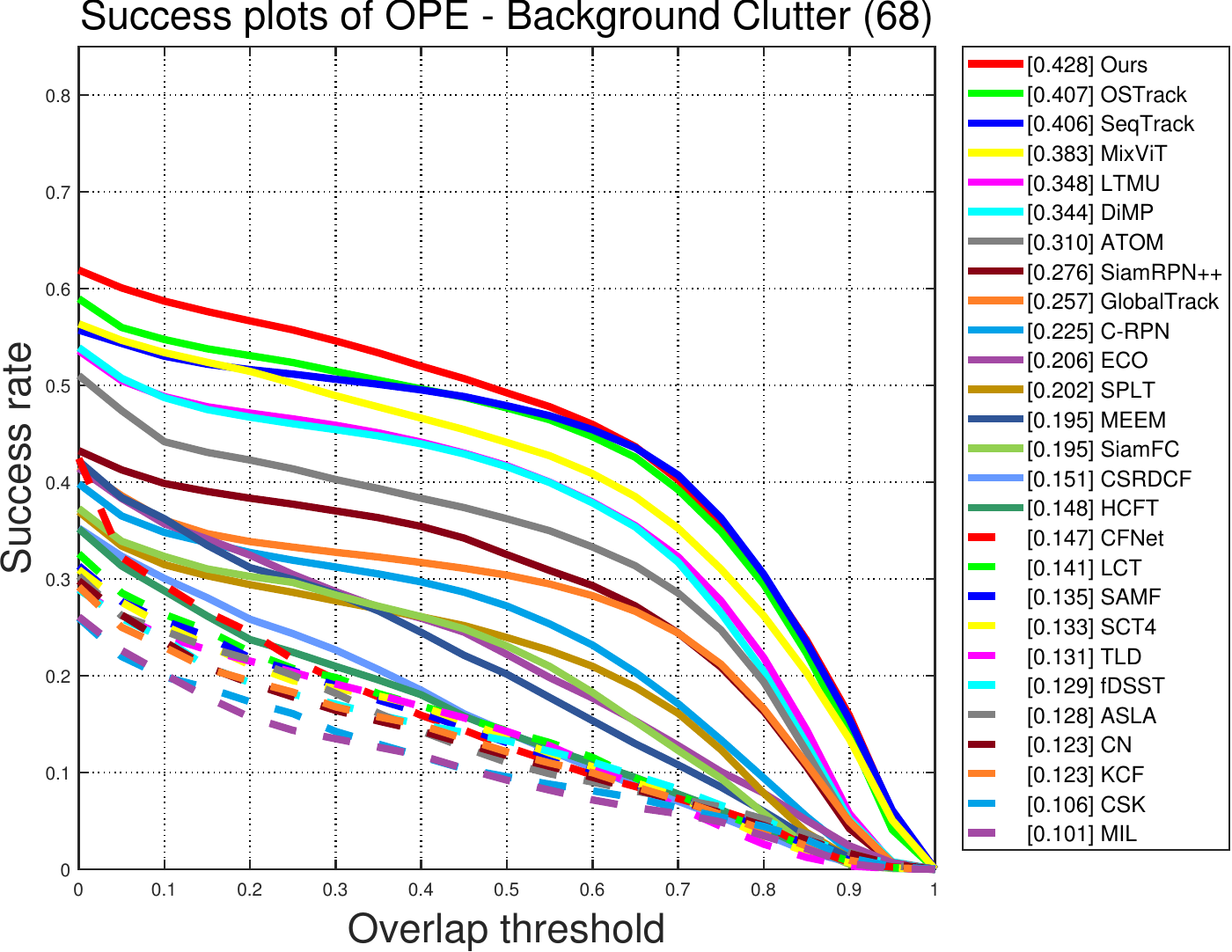}}
    \vspace{5pt}
	\centerline{\includegraphics[width=\textwidth]{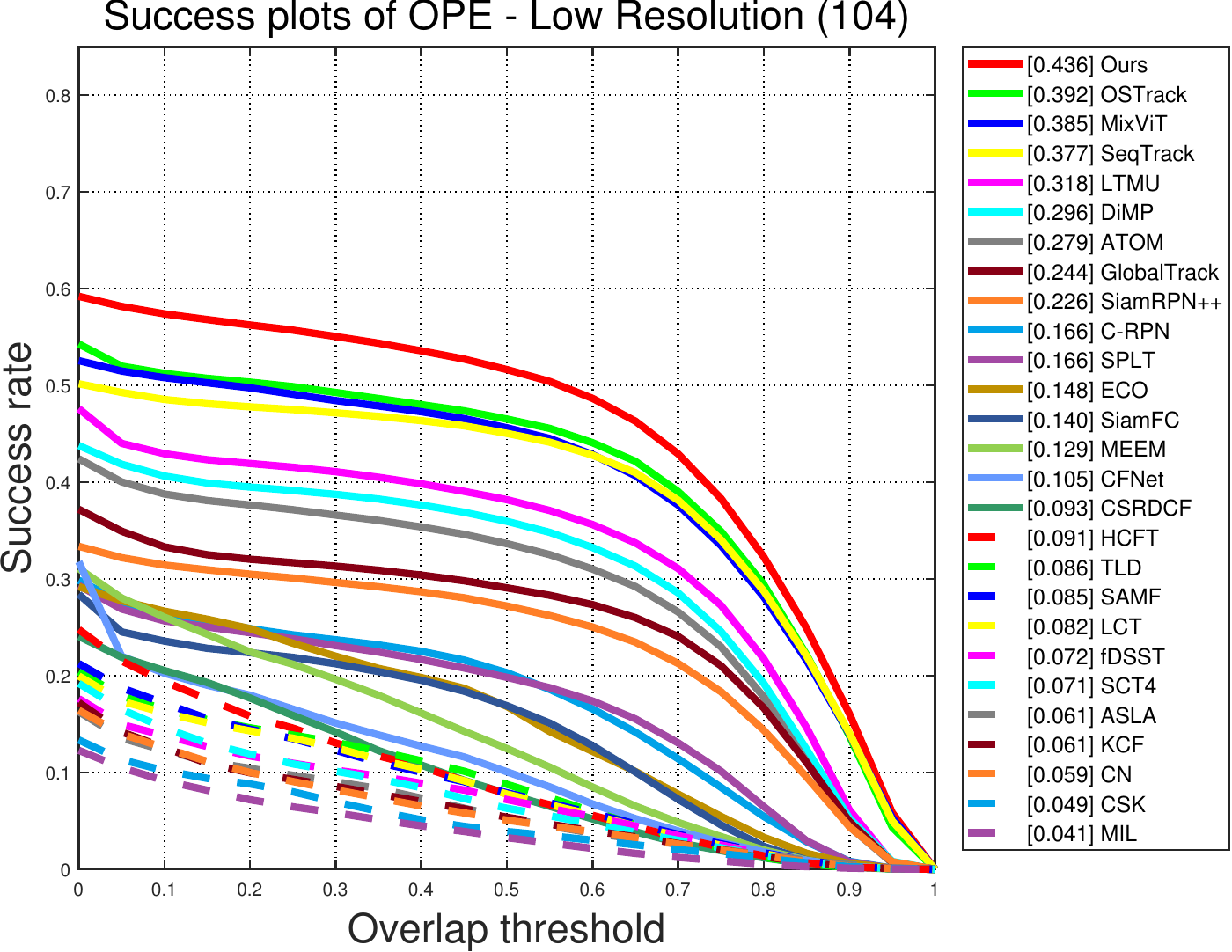}}
	\vspace{5pt}
	\centerline{\includegraphics[width=\textwidth]{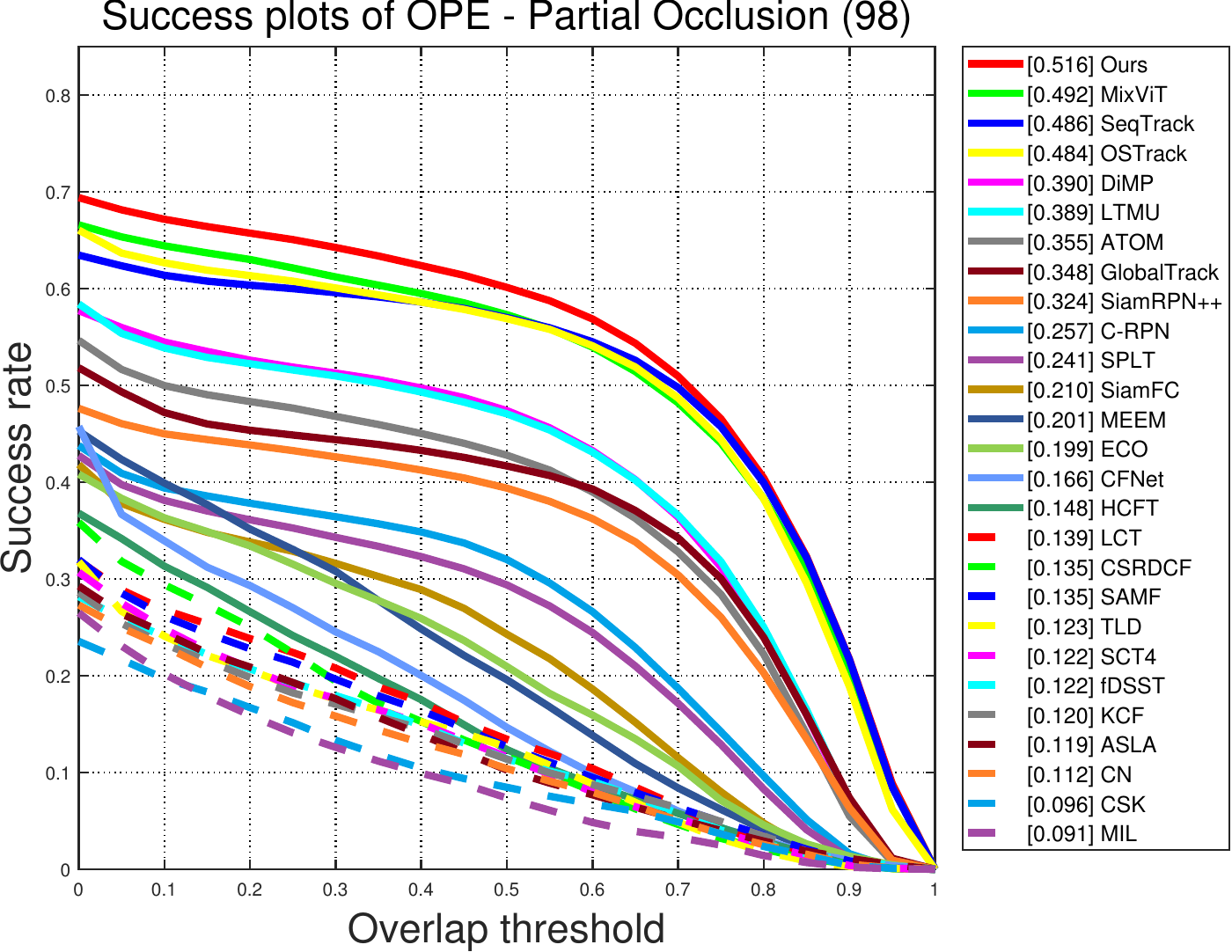}}
\end{minipage}
\caption{\textbf{Success plots of different attributes on LaSOT$_{ext}$.}}
\label{fig:lasot_ext-1}
\end{figure*}

\begin{figure*}[!t]
 \begin{minipage}{0.48 \linewidth}
 	\vspace{5pt}
 	\centerline{\includegraphics[width=\textwidth]{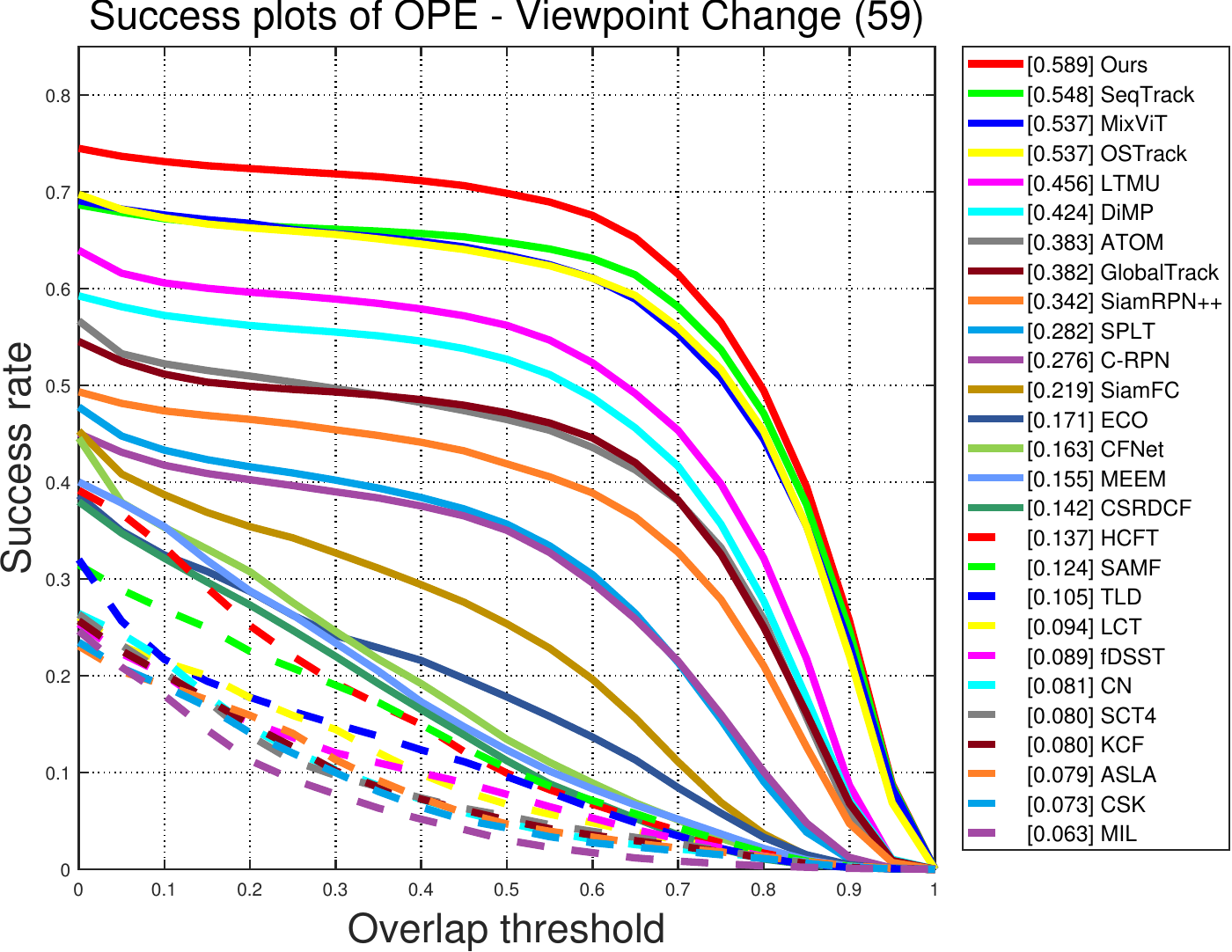}}
  	\vspace{5pt}
	  \centerline{\includegraphics[width=\textwidth]{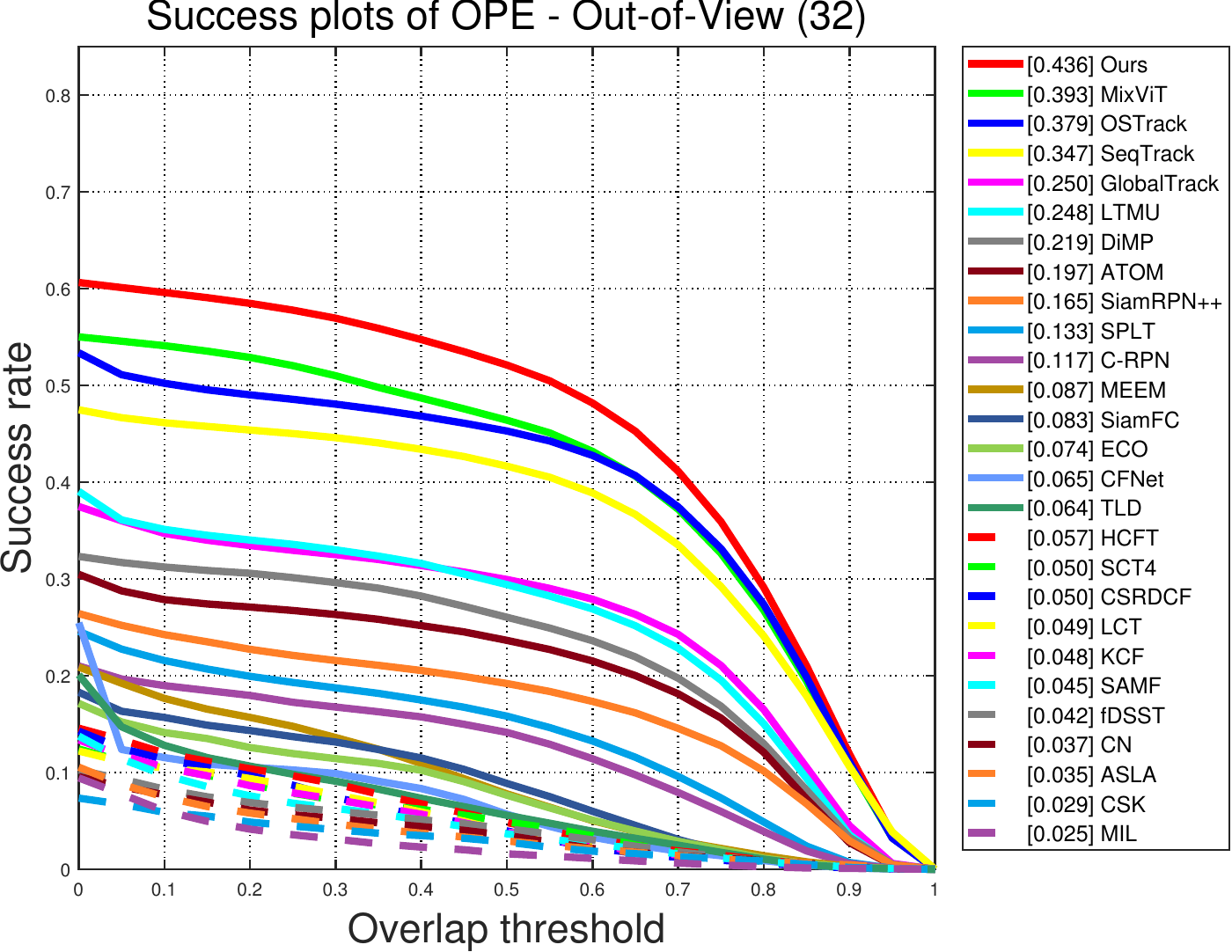}}
	  \vspace{5pt}
	  \centerline{\includegraphics[width=\textwidth]{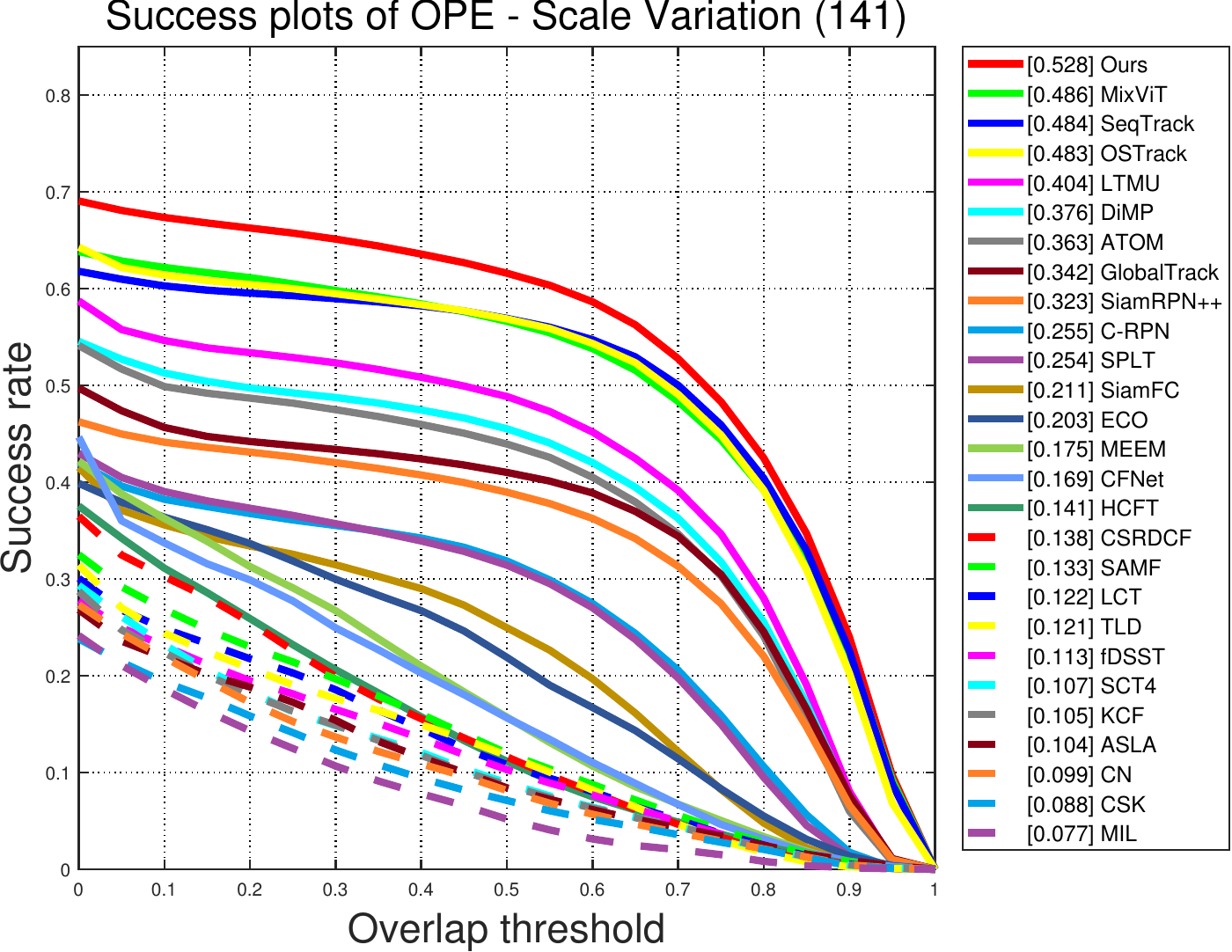}}
 \end{minipage}
 \begin{minipage}{0.48\linewidth}
	\vspace{5pt}
	\centerline{\includegraphics[width=\textwidth]{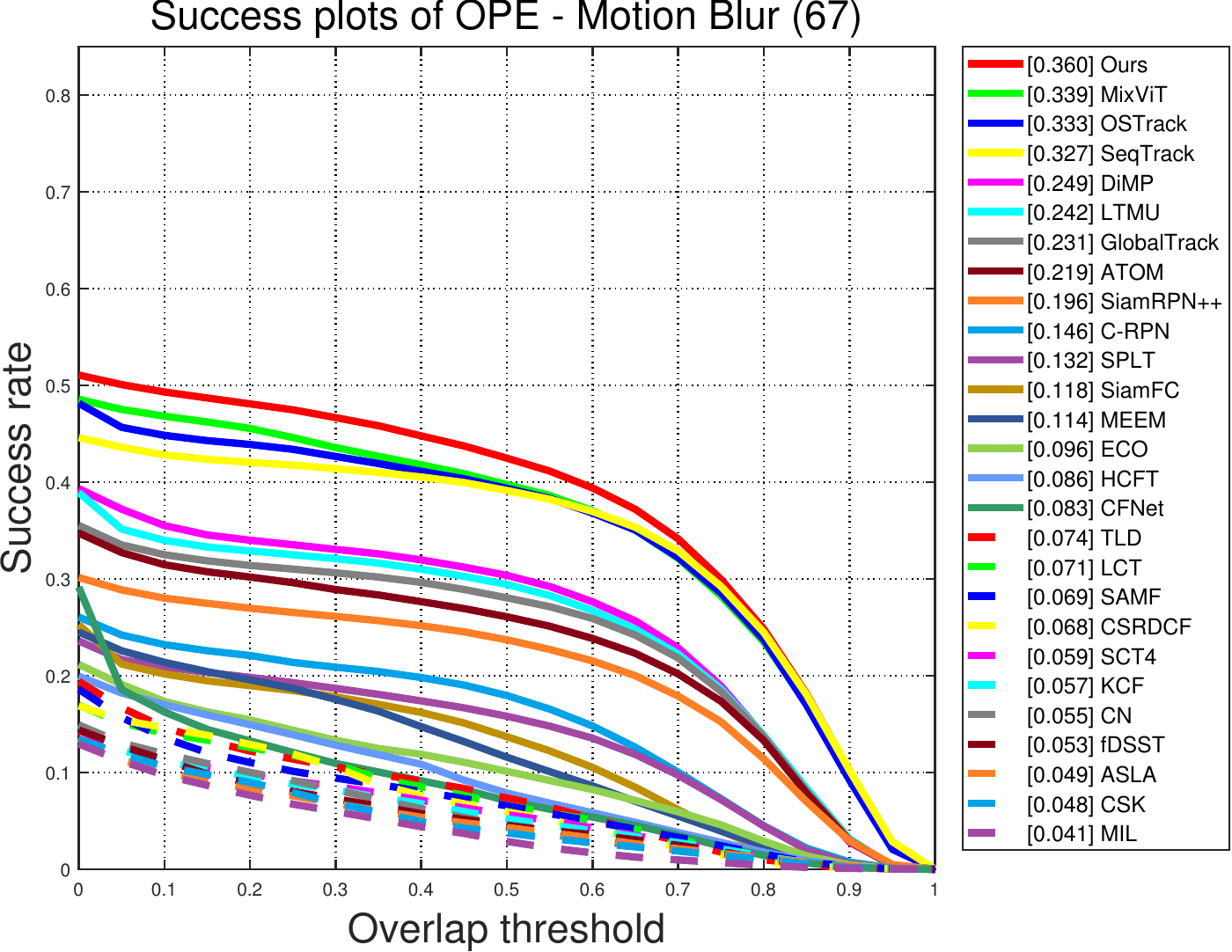}}
    \vspace{5pt}
	\centerline{\includegraphics[width=\textwidth]{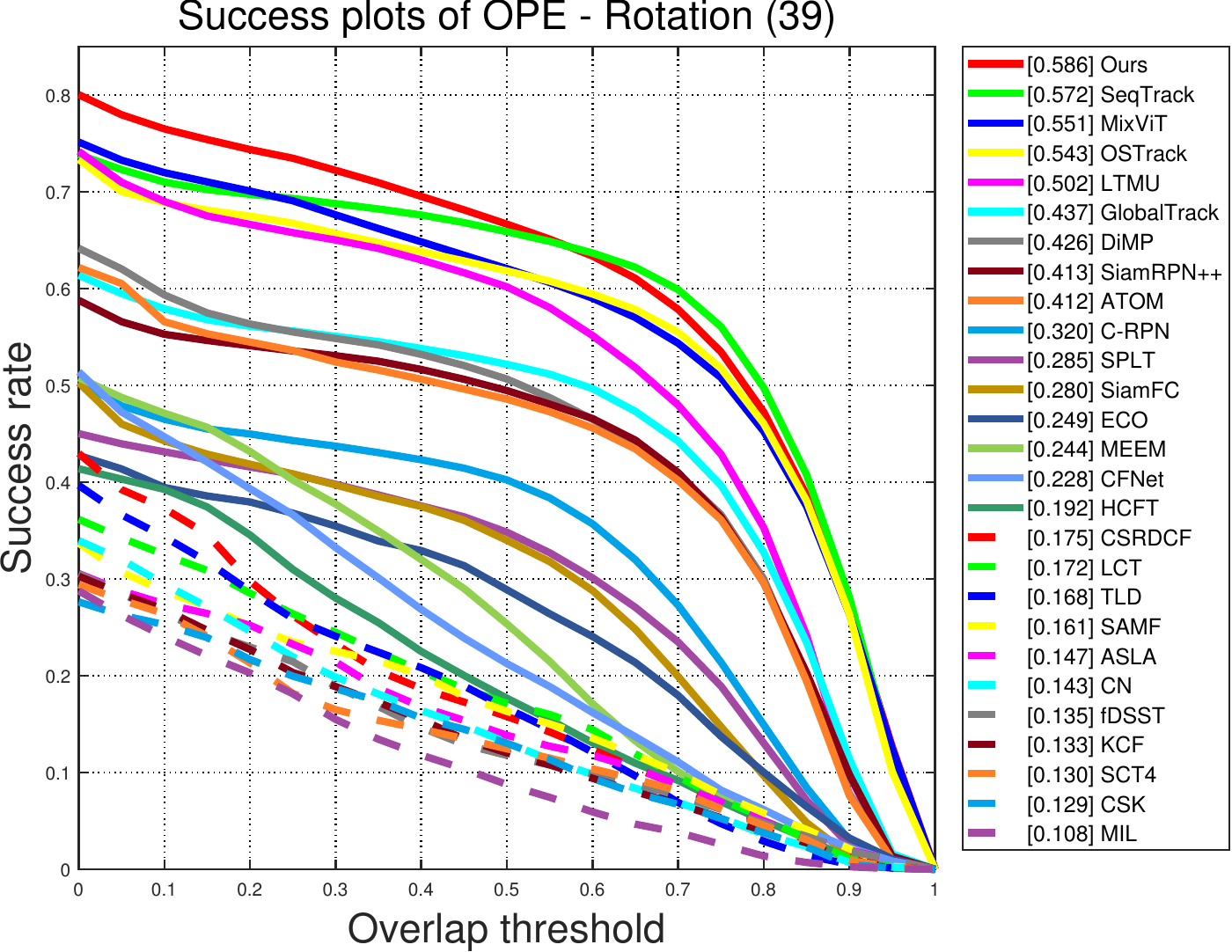}}
	\vspace{5pt}
	\centerline{\includegraphics[width=\textwidth]{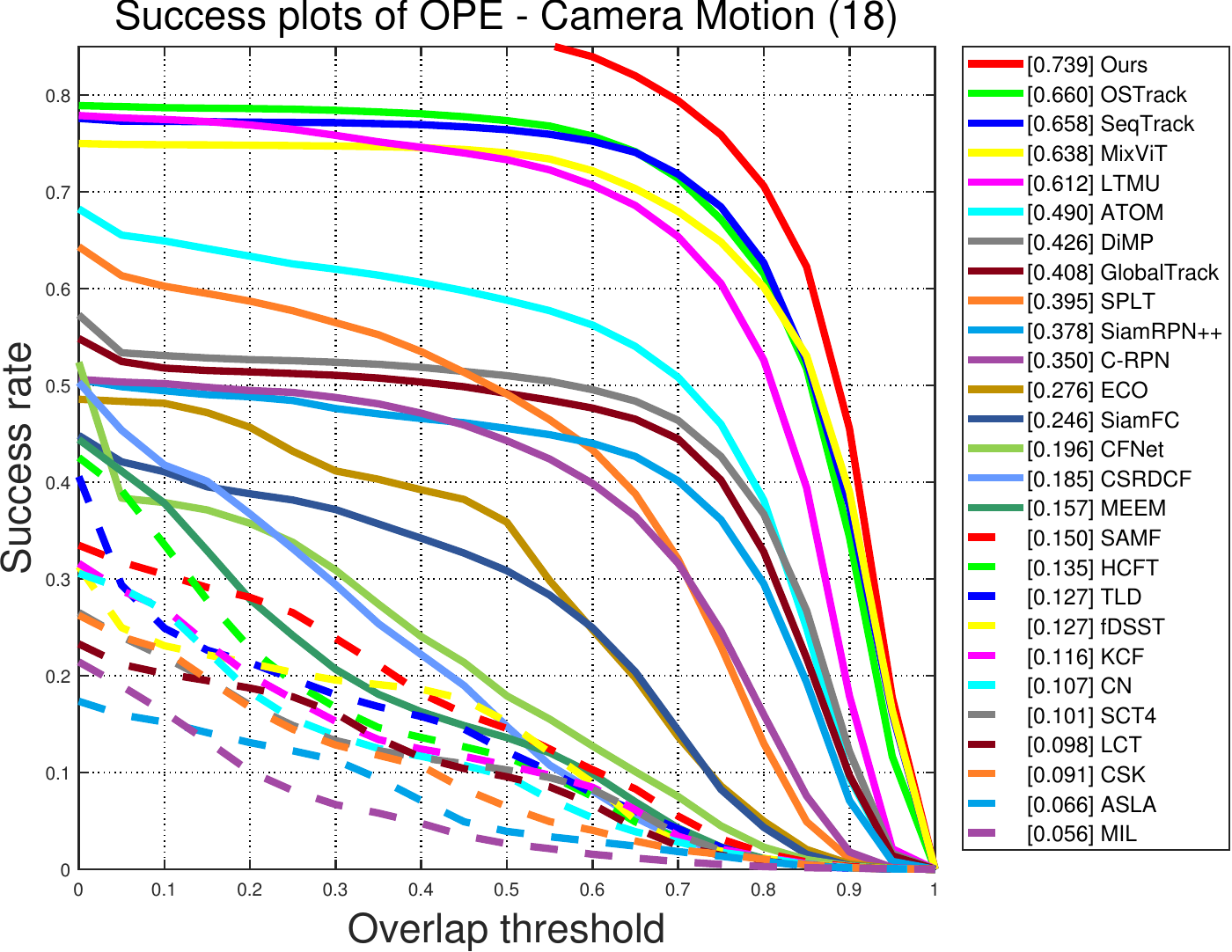}}
\end{minipage}
\caption{\textbf{Success plots of different attributes on LaSOT$_{ext}$.}}
\label{fig:lasot_ext-2}
\end{figure*}

The AUC score under different attributes of LaSOT$_{ext}$~\cite{LaSOT_ext} test set is shown in Figure~\ref{fig:lasot_ext-1} and Figure~\ref{fig:lasot_ext-2}.
It is worth noting that compared to the second-best tracker, our proposed method achieves an improvement of 5.1\% and 4.4\% AUC scores in fast-motion and low-resolution cases on the LaSOT$_{ext}$ test set, which benefits from the effective aid of the detection.
\vspace{-2mm}
\subsection{Detailed results on TNL2K}
\begin{figure*}[!t]
 \begin{minipage}{0.48 \linewidth}
 	\vspace{5pt}
 	\centerline{\includegraphics[width=\textwidth]{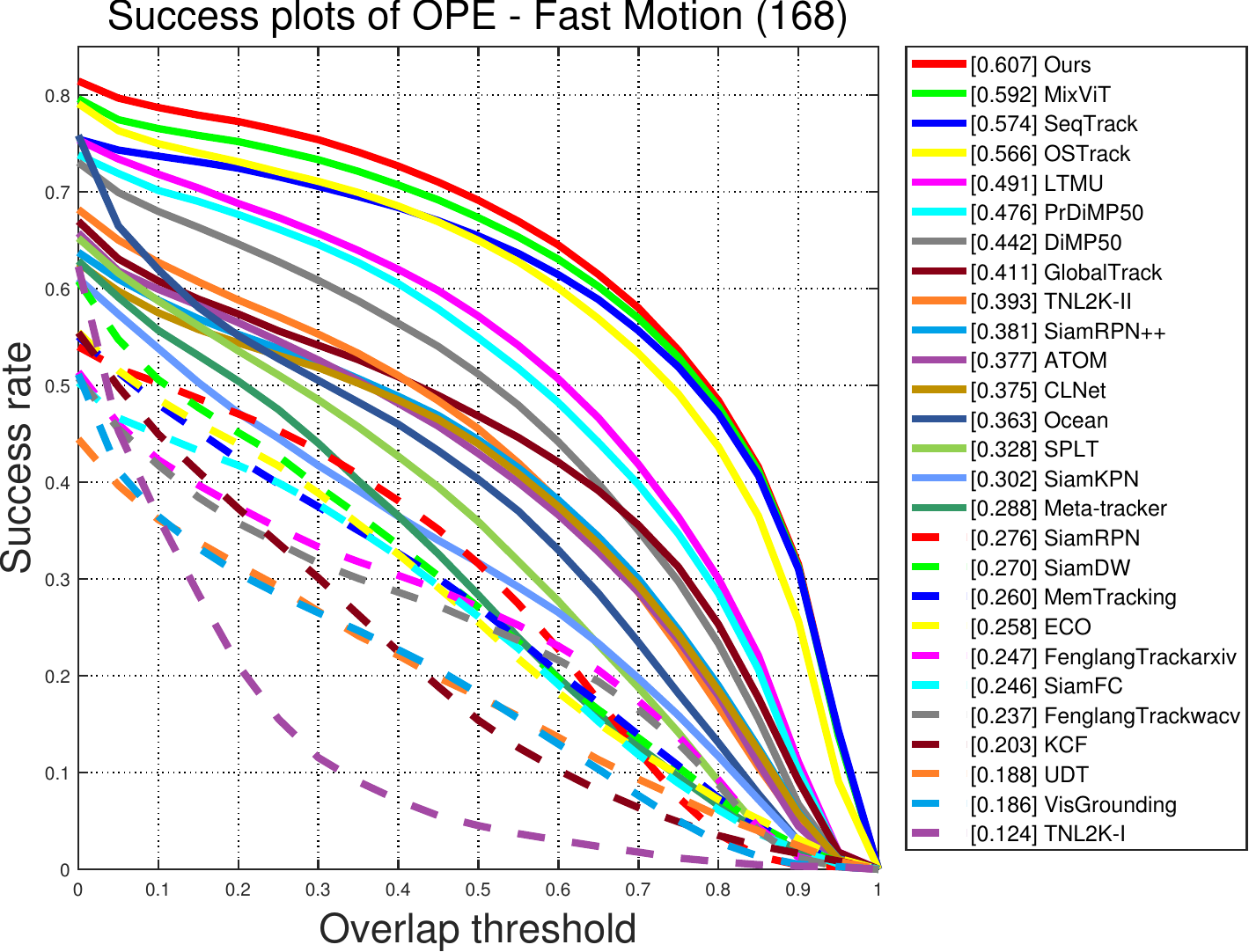}}
  	\vspace{5pt}
	  \centerline{\includegraphics[width=\textwidth]{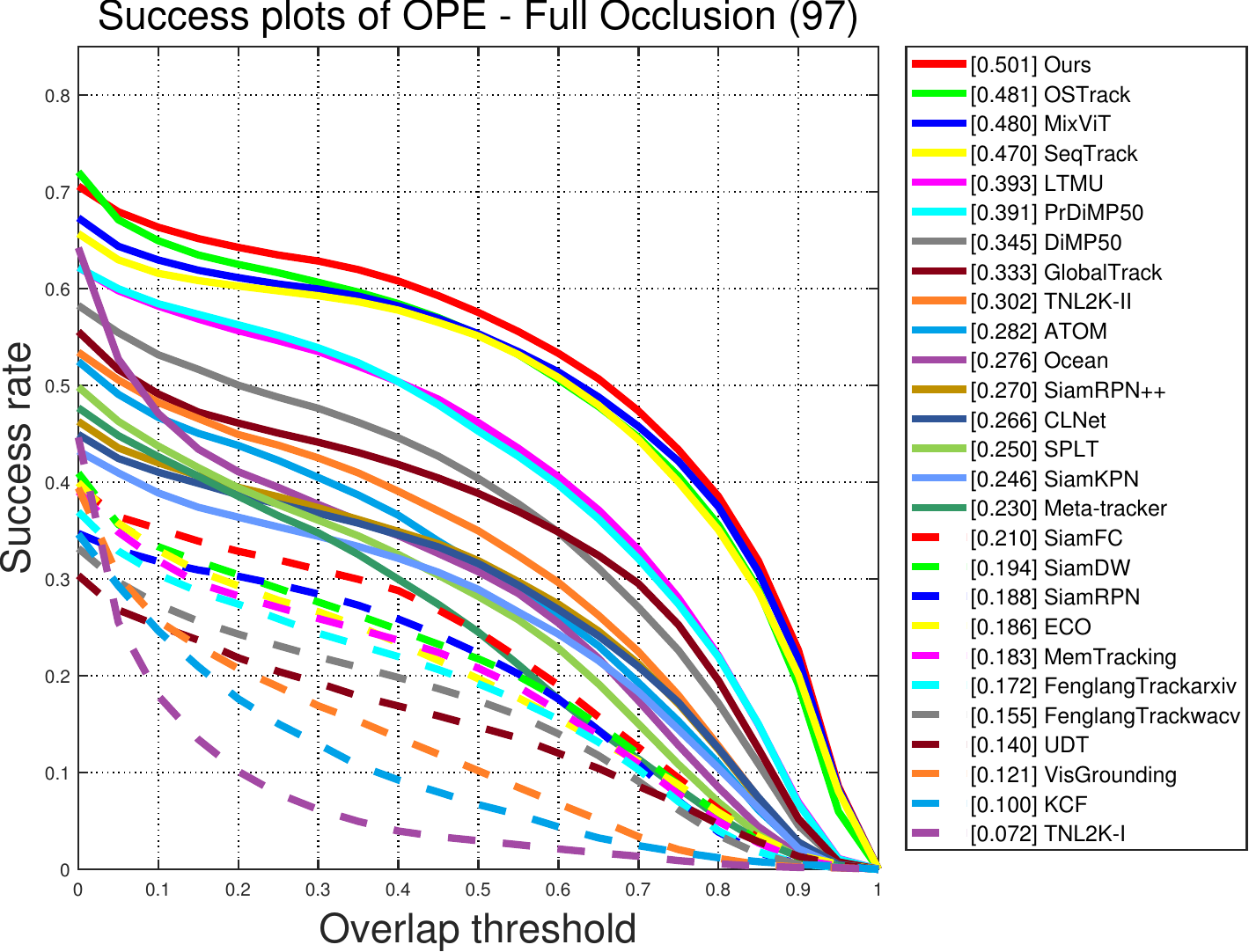}}
	  \vspace{5pt}
	  \centerline{\includegraphics[width=\textwidth]{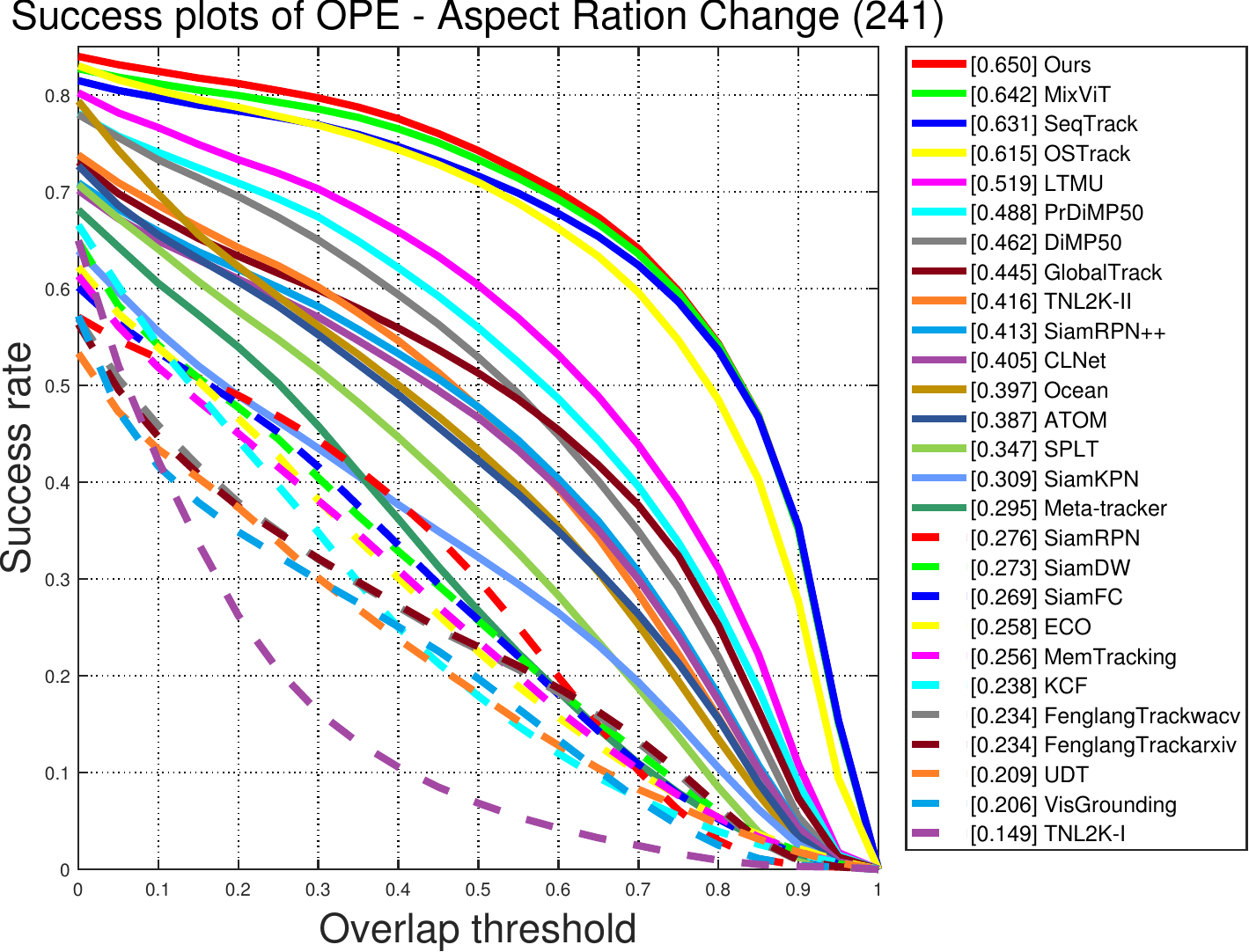}}
 \end{minipage}
 \begin{minipage}{0.48\linewidth}
	\vspace{5pt}
	\centerline{\includegraphics[width=\textwidth]{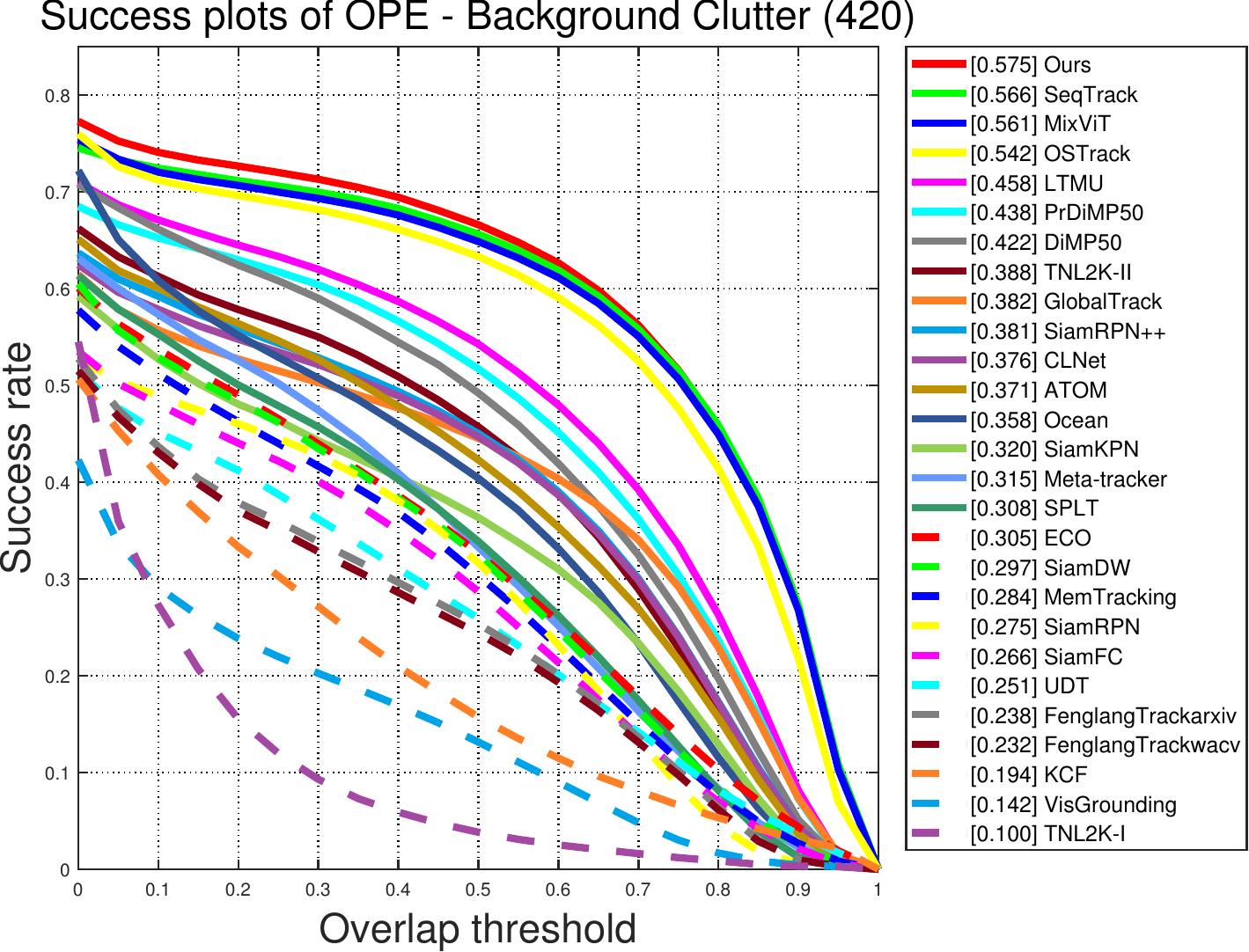}}
    \vspace{5pt}
	\centerline{\includegraphics[width=\textwidth]{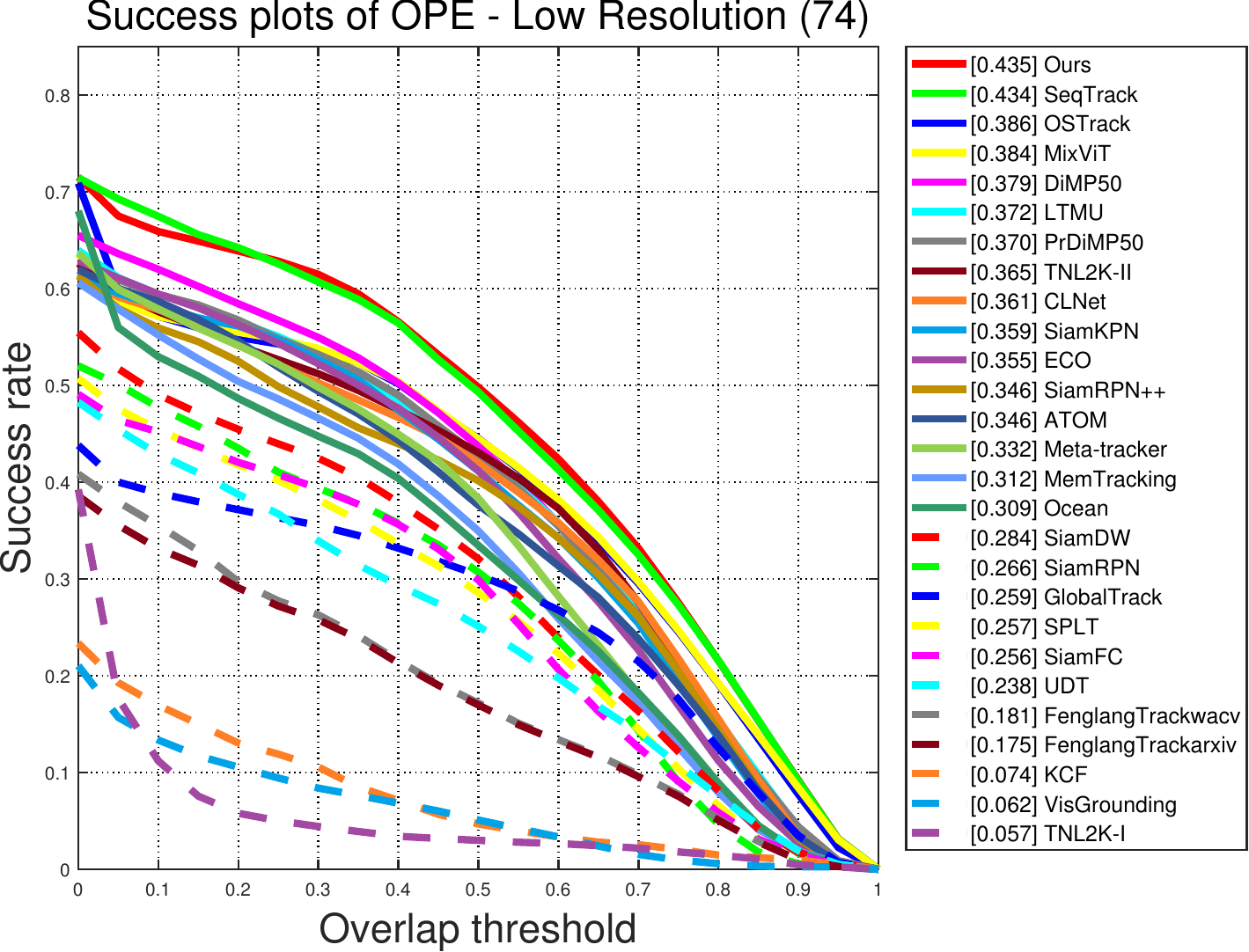}}
	\vspace{5pt}
	\centerline{\includegraphics[width=\textwidth]{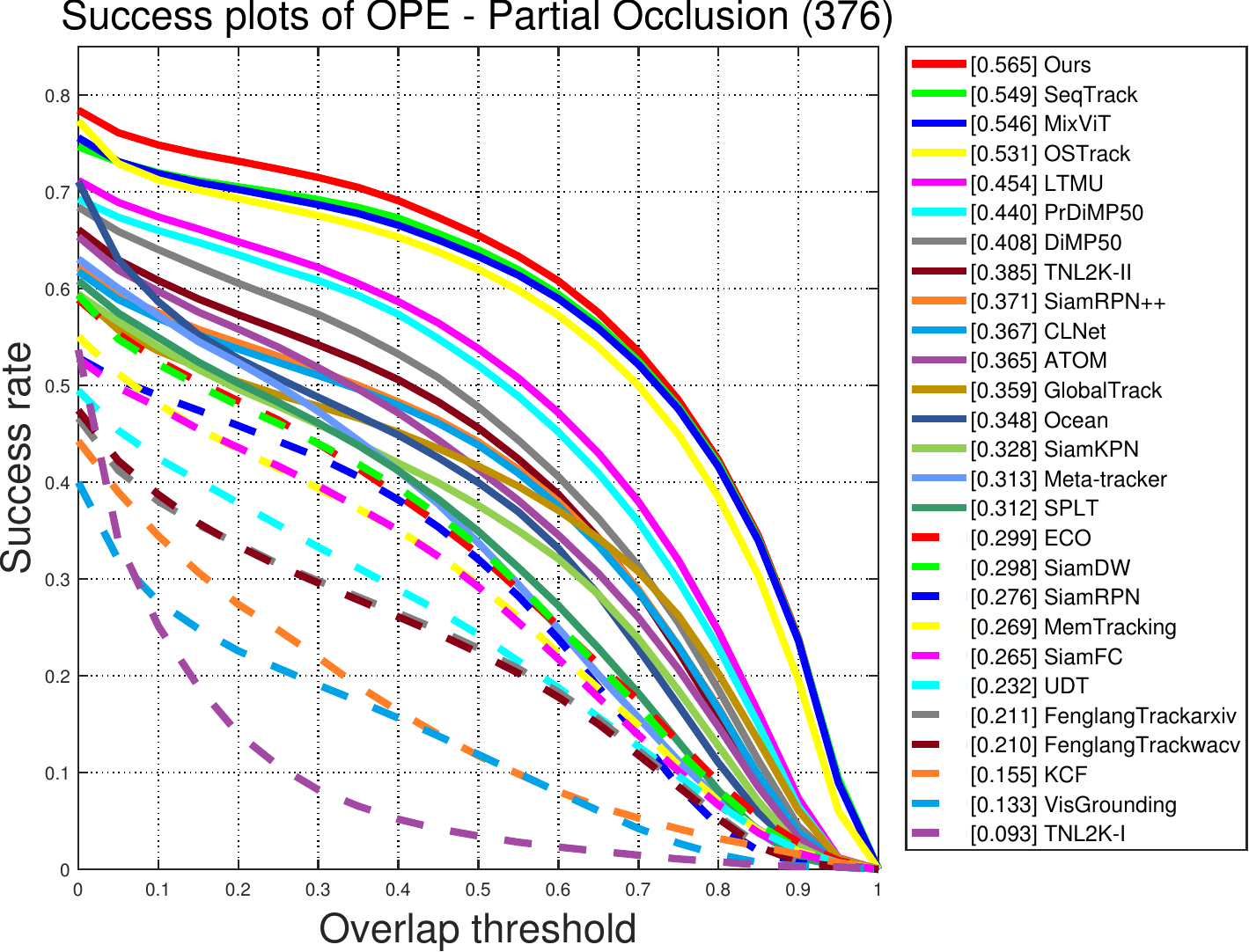}}
\end{minipage}
\caption{\textbf{Success plots of different attributes on TNL2K.}}
\label{fig:tnl2k-1}
\end{figure*}

\begin{figure*}[!t]
 \begin{minipage}{0.48 \linewidth}
 	\vspace{5pt}
 	\centerline{\includegraphics[width=\textwidth]{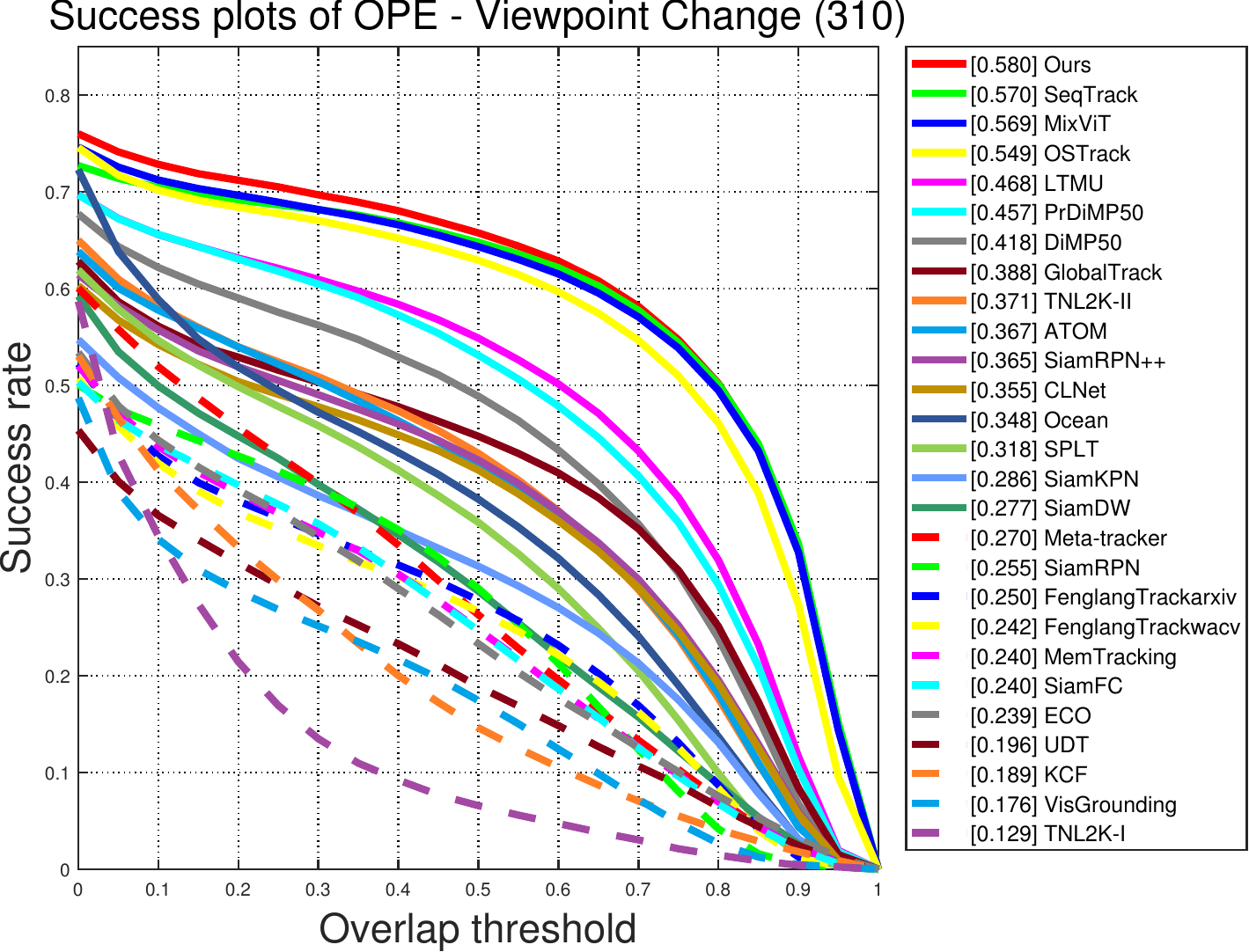}}
  	\vspace{5pt}
	  \centerline{\includegraphics[width=\textwidth]{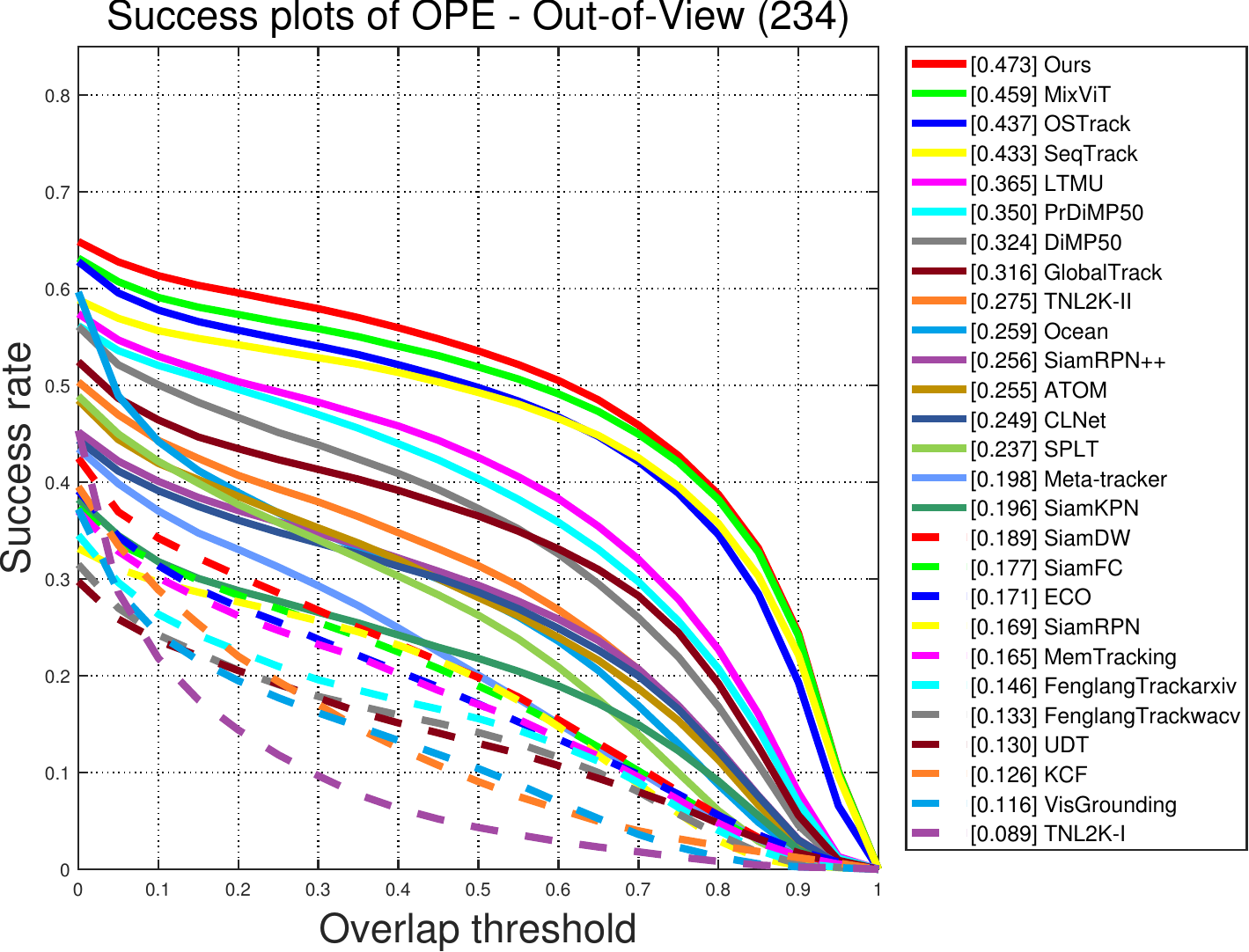}}
	  \vspace{5pt}
	  \centerline{\includegraphics[width=\textwidth]{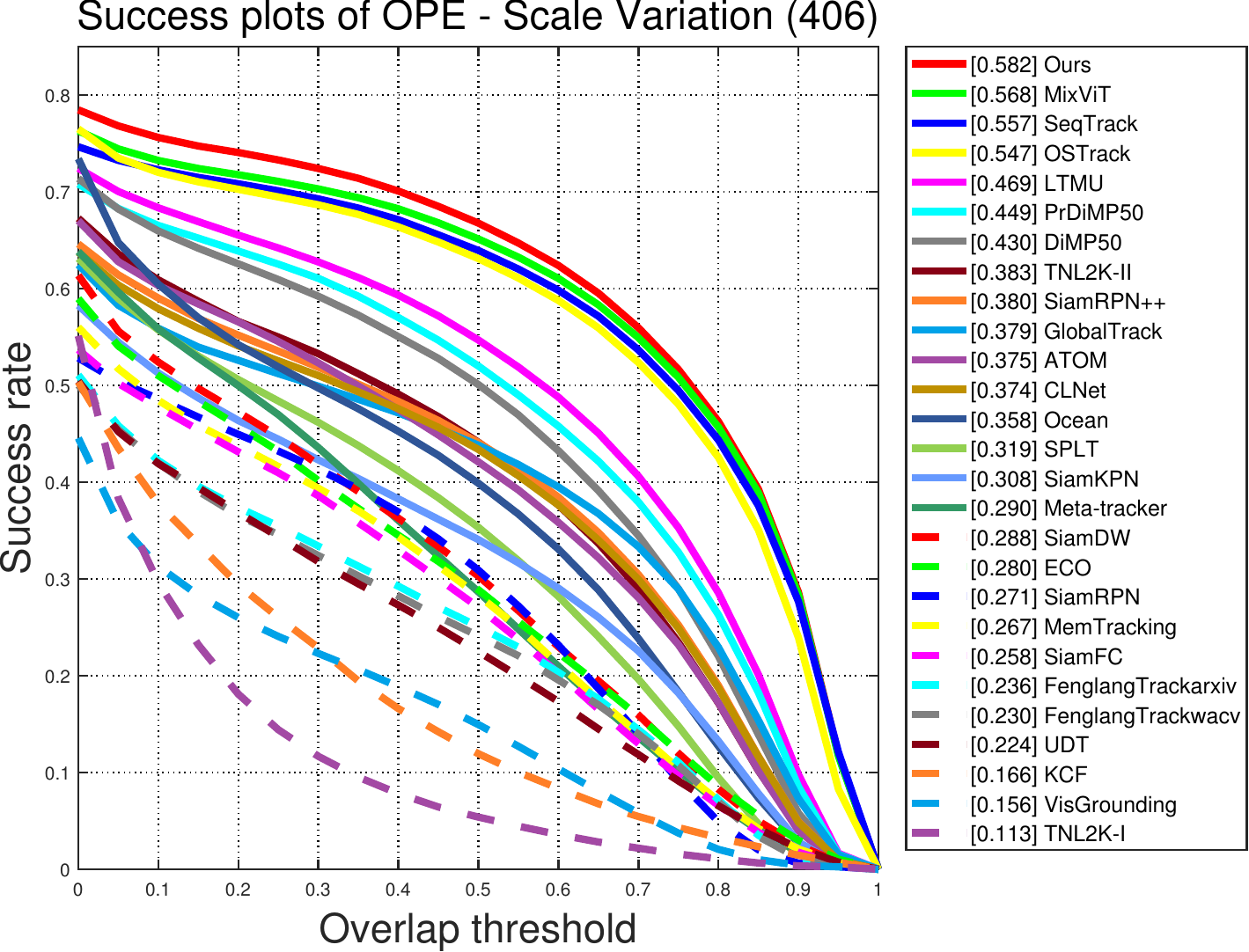}}
 \end{minipage}
 \begin{minipage}{0.48\linewidth}
	\vspace{5pt}
	\centerline{\includegraphics[width=\textwidth]{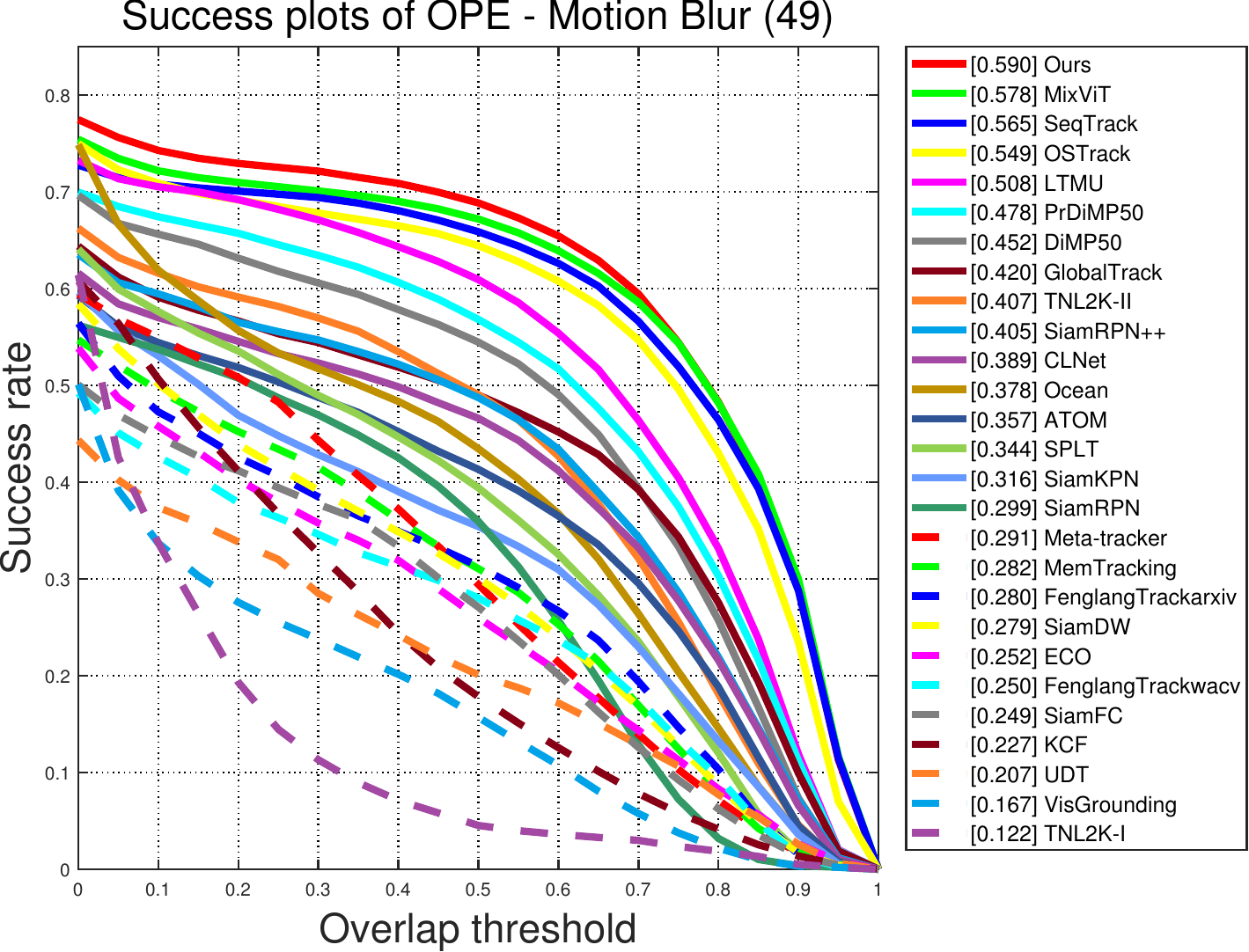}}
    \vspace{5pt}
	\centerline{\includegraphics[width=\textwidth]{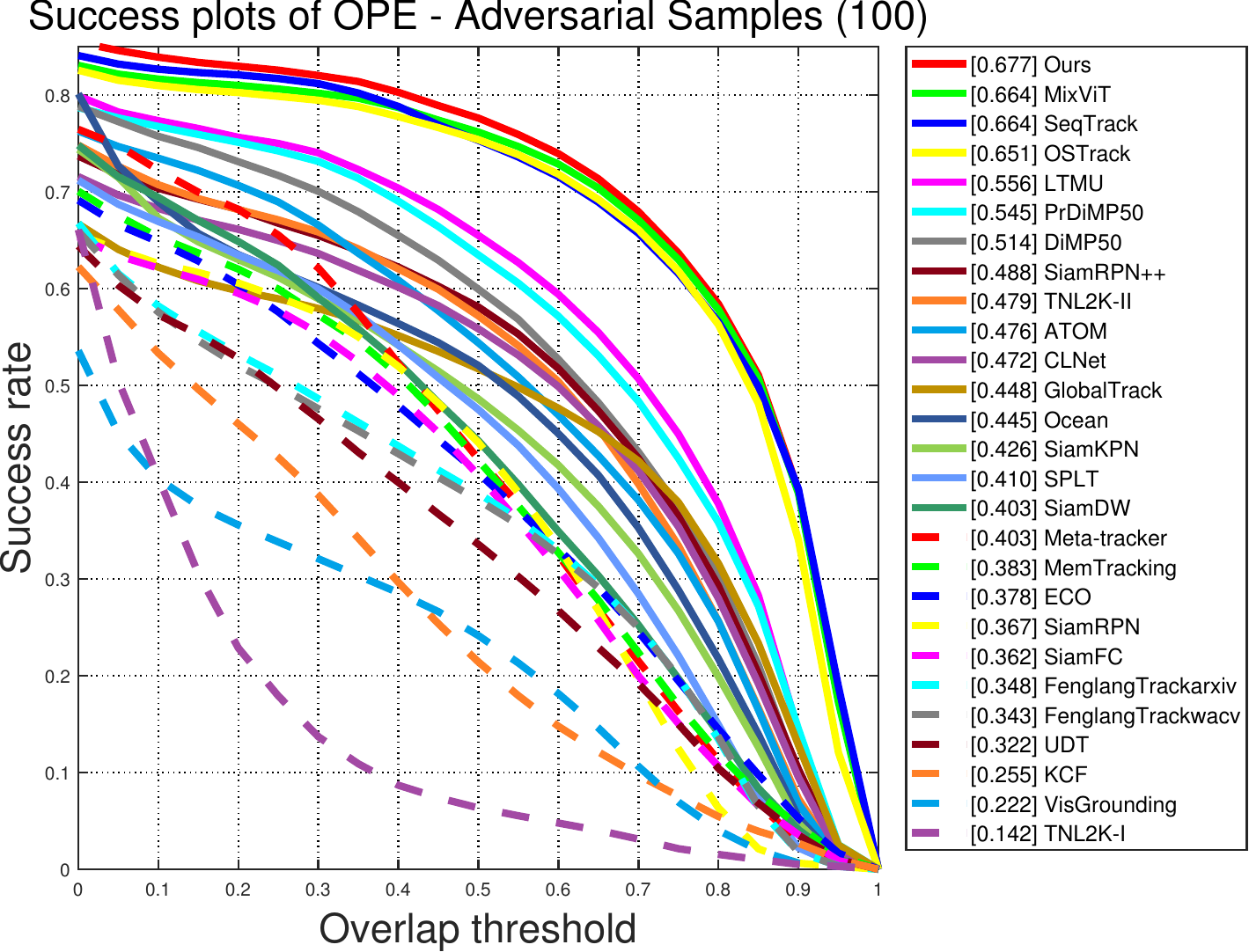}}
	\vspace{5pt}
	\centerline{\includegraphics[width=\textwidth]{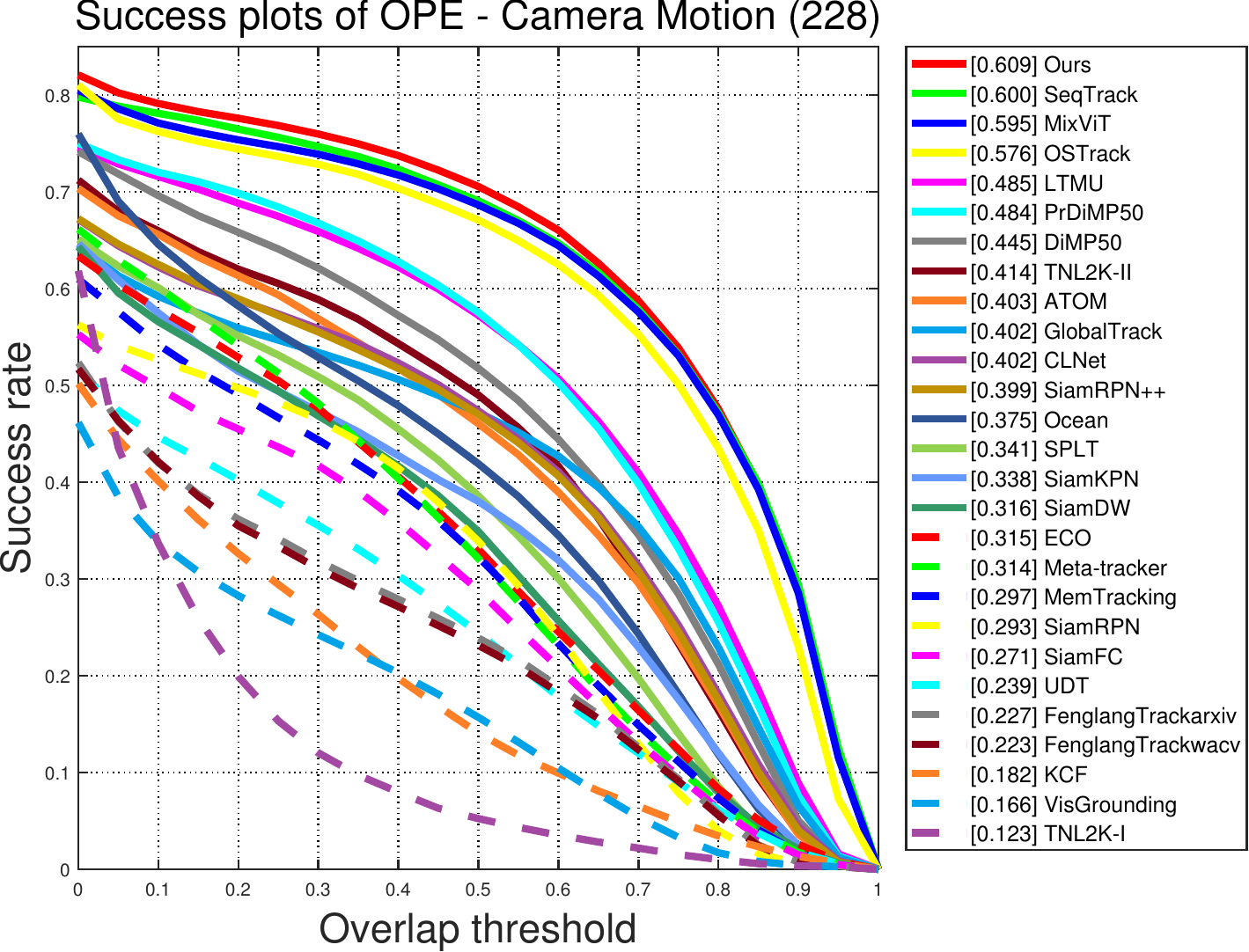}}
\end{minipage}
\caption{\textbf{Success plots of different attributes on TNL2K.}}
\label{fig:tnl2k-2}
\end{figure*}

The AUC score under different attributes of TNL2K~\cite{TNL2K} test set is shown in Figure~\ref{fig:tnl2k-1} and Figure~\ref{fig:tnl2k-2}.
TNL2K has introduced a novel challenge involving adversarial samples.
Compared to other trackers, our proposed method shows a 1.3\% increase in AUC scores, highlighting the effectiveness of our target state prediction in distinguishing the target from similar objects.


